\documentclass[10pt,twocolumn,letterpaper]{article}

\usepackage{iccv}
\usepackage{times}
\usepackage{epsfig}
\usepackage{graphicx}
\usepackage{amsmath}
\usepackage{amssymb}

\usepackage{subcaption}
\usepackage{multirow}
\usepackage{pifont}

\usepackage{algorithm}
\usepackage{algpseudocode}
\usepackage{stfloats}
\usepackage{appendix}

\usepackage[pagebackref=true,breaklinks=true,letterpaper=true,colorlinks,bookmarks=false]{hyperref}

\iccvfinalcopy % *** Uncomment this line for the final submission

\ificcvfinal\fi

\usepackage{xcolor}

\begin{document}

%%%%%%%%% TITLE
%\title{Boosting Adversarial Attacks through Enhanced Momentum}
\title{Boosting Adversarial Transferability through Enhanced Momentum}
%\title{An Enhanced Momentum Approach to Boosting Adversarial Transferability}

\author{Xiaosen Wang\textsuperscript{\rm 1}\thanks{The first two authors contribute equally.} \quad Jiadong Lin\textsuperscript{\rm 1}\footnotemark[1] \quad Han Hu\textsuperscript{\rm 2} \quad Jingdong Wang\textsuperscript{\rm 2} \quad Kun He\textsuperscript{\rm 1}\thanks{Corresponding author.}\\
\textsuperscript{\rm 1}School of Computer Science and Technology, Huazhong University of Science and Technology\\
\textsuperscript{\rm 2}Microsoft Research Asia\\
{\tt\small \{xiaosen,jdlin\}@hust.edu.cn, \{hanhu,jingdw\}@microsoft.com, brooklet60@hust.edu.cn}
}
% \and
%\author{Xiaosen Wang\textsuperscript{\rm 1}\thanks{The first two authors contribute equally.} \quad Jiadong Lin\textsuperscript{\rm 1}\footnotemark[1] \quad Han Hu\textsuperscript{\rm 2} \quad Jingdong Wang\textsuperscript{\rm 2} \quad Kun He\textsuperscript{\rm 1}\thanks{Corresponding author.}\\
%\textsuperscript{\rm 1}School of Computer Science and Technology, Huazhong University of Science and Technology\\
%\textsuperscript{\rm 2}Microsoft Research Asia\\
%{\tt\small \{xiaosen,jdlin\}@hust.edu.cn, \{hanhu,jingdw\}@microsoft.com, brooklet60@hust.edu.cn}

\maketitle
\ificcvfinal\thispagestyle{empty}\fi

%%%%%%%%% ABSTRACT
\begin{abstract}
Deep learning models are known to be vulnerable to adversarial examples crafted by adding human-imperceptible perturbations on benign images. Many existing adversarial attack methods have achieved great white-box attack performance, but exhibit low transferability when attacking other models. Various momentum iterative gradient-based methods are shown to be effective to improve the adversarial transferability. 
% As the general optimization methods, however, they did not consider the neighborhood constraint of the adversaries, which should be considered in the adversarial learning scenario.
%, leading to limited improvement. In this work,
% To this end, 
% we propose an enhanced momentum iterative gradient-based method that considers the neighborhood constraint of the crafted examples. 
%to further improve the transferability. 
%In this work
In what follows, we propose an enhanced momentum iterative gradient-based method to further enhance the adversarial transferability. 
Specifically, instead of only accumulating the gradient during the iterative process, we additionally accumulate the average gradient of the data points sampled %from a line segment determined by 
in the gradient direction of the previous iteration 
so as to stabilize the update direction and escape from poor local maxima.
Extensive experiments on the standard ImageNet dataset demonstrate that our method could improve the adversarial transferability of momentum-based methods by a large margin of 11.1\% on average. 
Moreover, by incorporating with various input transformation methods, the adversarial transferability 
could be further improved significantly. 
We also attack several extra advanced defense models under the ensemble-model setting, and the enhancements are remarkable with at least 7.8\% on average. %, which further validate the effectiveness of the enhanced momentum approach.
% Our anonymous code is publicly available at https: ...\HK{add}

% Deep learning models are known to be vulnerable to adversarial examples crafted by adding human-imperceptible perturbations on benign images. Many existing adversarial attack methods achieve great white-box attack performance, but exhibit low transferability when attacking other models. Various momentum iterative gradient-based methods are shown to be effective to improve the adversarial transferability. In this work, we propose an enhanced momentum iterative gradient-based method to further enhance the adversarial transferability. Specifically, instead of only accumulating the gradient during the iterative process, we additionally accumulate the average gradient of the data points sampled in the gradient direction of the previous iteration so as to stabilize the update direction and escape from poor local maxima. Extensive experiments on the standard ImageNet dataset demonstrate that our method could improve the adversarial transferability of momentum-based methods by a large margin of 11.1\% on average. Moreover, by incorporating with various input transformation methods, the adversarial transferability could be further improved significantly. We also attack several extra advanced defense models under the ensemble-model setting, and the enhancements are remarkable with at least 7.8\% on average.
\end{abstract}

%%%%%%%%% BODY TEXT
\begin{figure}
    \centering
    % \vspace{-0.5em}
    \includegraphics[width=\linewidth]{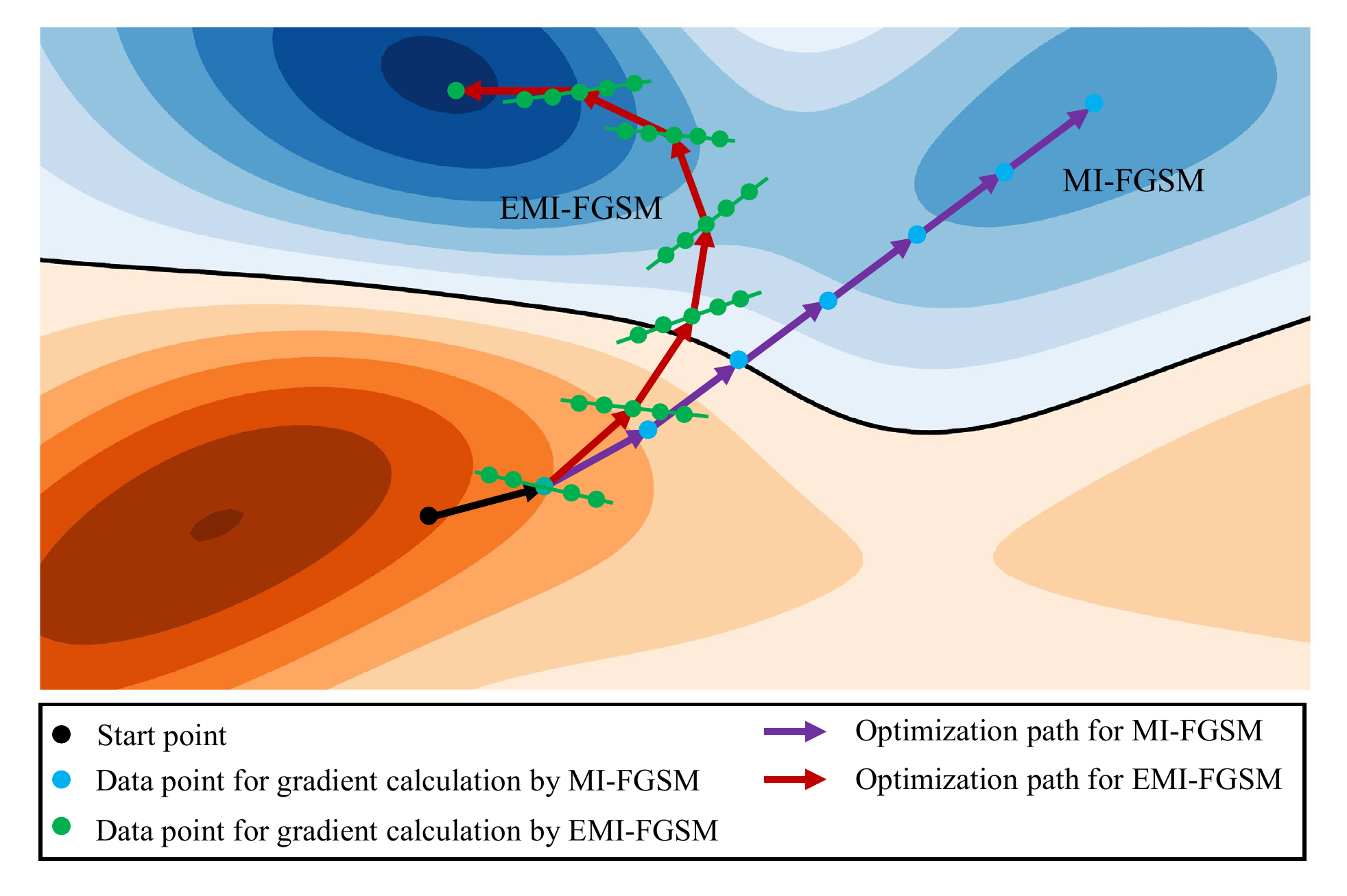}
    \caption{Illustration of the optimization path of MI-FGSM~\cite{dong2018boosting} and the proposed EMI-FGSM. At each iteration, MI-FGSM accumulates the gradient of data point along the path, while EMI-FGSM accumulates the accumulated gradient of sampled data points in the gradient direction of previous iteration. With such accumulation, EMI-FGSM can find better local maxima and exhibits higher transferability.}
    % , where $\bar{g}_{t-1}$ is the average gradient of the previous iteration.}
    \label{fig:optimization}
    % \vspace{-1em}
\end{figure}
\section{Introduction}

With the impressive performance of deep neural networks (DNNs) \cite{krizhevsky2012imagenet,he2016resnet,girshick2015fast,long2015fully,devlin2019bert}, the vulnerability to adversarial examples \cite{szegedy2014intriguing, goodfellow2015FGSM}, which are indistinguishable from legitimate ones by adding tiny perturbations but lead to erroneous predictions, has raised serious concerns in security-sensitive applications, \eg self-driving automobile~\cite{eykholt2018robust}, face verification~\cite{sharif2016accessorize} \etc. This issue of DNNs has triggered two research directions, with one trying to improve the attack ability of adversarial examples \cite{goodfellow2015FGSM,kurakin2017IFGSM,carlini2017towards,athalye2018obfuscated,li2019nattack} and the other line studying to improve the robustness of neural networks against the adversaries \cite{madry2018pgd,zhang2019theoretically,liao2018defense,xie2018mitigating,cohen2019certified}. The two directions, namely \emph{adversarial attack} and \emph{adversarial defense}, usually act like spear and shield that the progress on one side can inspire the improvements of the other side.
% \han{This issue of deep neural networks has triggered two research directions, with one trying to improve the attack ability of adversarial examples and the other line studying how to improve the robustness of neural networks to the adversaries. The two directions, named \emph{adversarial attack} and \emph{adversarial defense}, usually act like spear and shield that the progress on one side can inspire the improvements of the other side. This paper studies the adversarial attack methods. [han: I commented previous sentences as below. If the edit is not proper, please roll it back.]}

%One intriguing property of the adversarial examples is that they are \textit{transferable} \cite{papernot2017practical,liu2017delving}, \ie the adversarial examples generated on one model can still mislead other models with high probability. Such property makes it easy to attack deep neural models under black-box setting in practice.

For adversarial attack, numerous methods have been proposed in recent years,
% \han{For adversarial attack, recently, numerous adversarial attack methods have been proposed,}
% Recently, numerous adversarial attack methods have been proposed,
such as the one-step gradient-based attacks \cite{goodfellow2015FGSM, tramer2018ensemble}, iterative gradient-based attacks \cite{kurakin2017IFGSM,madry2018pgd}, and optimization-based attacks \cite{szegedy2014intriguing,carlini2017towards}. Existing adversarial attacks often fall into the category of white-box setting, where the adversary is capable to access all information about the target model. 
For the counterpart category of black-box attacks, adversarial transferability, \ie the ability of adversarial examples generated on one model to mislead other models, is an important metric. Such property makes it possible to attack deep neural models without knowing any inner working mechanism in practice.
Though white-box attacks achieve good attack performance, they often exhibit low transferability. %, \ie the ability of adversarial examples generated on one model to mislead other models. Such property makes it possible to attack deep neural models without knowing any inner working mechanism in practice.%under black-box setting in practice.
%The threat of adversarial examples has also inspired a broad range of adversarial defense methods, \eg adversarial training \cite{goodfellow2015FGSM,kurakin2017IFGSM,tramer2018ensemble,zhang2019theoretically, Ding2020MMA}, input preprocessing \cite{liao2018defense,xie2018mitigating,liu2019FD,naseer2020NRP}, and theoretically certified defense \cite{cohen2019certified, salman2019provably}.

Recently, various methods have been proposed to improve the transferability of white-box attacks, \eg incorporating momentum into iterative gradient-based attacks~\cite{dong2018boosting,lin2020nesterov}, ensemble-model attack~\cite{liu2017delving}, input transformations~\cite{xie2019improving,dong2019evading,lin2020nesterov,wang2021Admix} \etc. Note that both the ensemble-model attack and input transformations are based on existing gradient-based attacks. 
% Among which, NI-FGSM~\cite{lin2020nesterov} exhibits the best transferability among existing momentum based attacks, and the authors demonstrated that NI-FGSM can be integrated with advanced input transformations to further improve the transferability.
However, NI-FGSM, which exhibits the best transferability among existing momentum based attacks \cite{lin2020nesterov}, can only achieve the average attack success rate of less than $52\%$ under black-box setting, as shown in Table~\ref{tab:comparison_inc_v3}, indicating that the improvement of ensemble-model attack and transformation-based attack is rather limited. 

% In order to improve the transferability of white-box attacks, Dong \etal~\cite{dong2018boosting} propose MI-FGSM, which integrates momentum~\cite{polyak1964some} into the iterative gradient-based attack. Lin \etal \cite{lin2020nesterov}  propose NI-FGSM, which further enhances the attack transferability with Nesterov accelerated gradient (NAG)~\cite{Nesterov1983}. Liu \etal \cite{liu2017delving} demonstrate that the ensemble attack, which attacks multiple models simultaneously, can improve the transferability. Besides, recent works propose various input transformations \cite{xie2019improving, dong2019evading, lin2020nesterov} to boost the transferability. Note that both the ensemble-model attack and input transformations are based on existing gradient-based attacks. However, NI-FGSM, which exhibits the best transferability among existing gradient-based attacks \cite{lin2020nesterov}, can only achieve the average attack success rate of less than $52\%$ under black-box setting, as shown in Table~\ref{tab:comparison_inc_v3}, indicating that the improvement of ensemble-model attack and transformation-based attack is rather limited. 

In this work, inspired by the current momentum based attacks, we propose 
%a class of \textit{enhanced momentum iterative gradient-based methods} to improve the attack success rates of the crafted adversarial examples, and promote the transferability
an enhanced momentum iterative fast gradient sign method, termed EMI-FGSM, to further promote the adversarial transferability. As shown in Figure~\ref{fig:optimization}, different from  the existing momentum based methods (\eg MI-FGSM) that just accumulate the gradients of data points along the optimization path, our enhanced momentum based method additionally accumulates the gradients of data points sampled in the gradient direction of previous iteration. Such accumulation might help find more stable direction of the gradient, leading to better local maxima. Empirical evaluations show that our method achieves higher attack success rates under white-box setting and exhibit significantly higher transferability under black-box setting. 
% We also visualize the adversaries crafted by various attacks in Figure \ref{fig:adv_images}, in which the proposed EMI-FGSM crafts visually similar adversaries but with much higher transferability as demonstrated in the experiments.

Moreover, the proposed EMI-FGSM approach can work complementary to ensemble-model attacks and various input transformations. When integrated with these advanced methods, the enhanced momentum equipped methods can achieve significantly higher transferability on the standard ImageNet dataset than the state-of-the-art baselines. When attacking seven advanced defense models that exhibit good defense effectiveness against transferability on ImageNet, our method combined with input transformations under ensemble-model setting achieves an average attack success rate of 86.6\%, improving the transferability of existing advanced methods by a clear margin of 7.8\%. 

% \han{In addition, the proposed approach, named EMI-FGSM, can work complementary to the methods based on ensemble-models and various input transformations. When combined with the advanced ensemble-model and input transformation based methods, it can achieve significantly greater transferability using ImageNet dataset. In attacking seven advanced defense methods that exhibit good defense efficacy against transferability on ImageNet, our methods combined with input transformations and ensemble models achieve an average attack success rate of 86.6\%, improving the transferability of existing state-of-the-art methods with a clear margin of 7.8\%.}

%To further improve the transferability of the adversarial attacks, we integrate our methods into the ensemble-model attack and various input transformations. Extensive experiments on ImageNet dataset demonstrate that the enhanced momentum-based methods can further improve the transferability with various input transformations under both single model and ensemble-model settings. 
%We also adopt seven advanced defense methods that exhibit good defense efficacy against transferability on ImageNet and our methods with input transformations under the ensemble-model setting achieve an average attack success rate of 86.6\%, improving the transferability of existing state-of-the-art methods with a clear margin of 7.8\%.

\section{Related Work}
Given a classifier $f$ and an input image $x$, where $f(x)$ outputs the prediction label of $x$. Let $J_f(x,y)$ denote the loss function of classifier $f$ and $\mathcal{B}_\epsilon(x) = \{x':\|x-x'\|_p \leq \epsilon\}$ denote the $L_p$-norm ball centered at $x$ with radius $\epsilon$ and we focus on $L_\infty$-norm as in previous works.

\subsection{Adversarial Attacks}
Adversarial attacks can be formulated as finding an example $x^{adv} \in \mathcal{B}_\epsilon(x)$ that satisfies $f(x) \neq f(x^{adv})$. According to the threat model, existing adversarial attacks can be roughly categorized into two settings: a) \textit{white-box attack} allows full access to the threat model, \eg model outputs, (hyper-)parameters, gradients and architectures, \etc b) \textit{black-box attack} only allows access to the model outputs. Recent works also find that adversarial examples have good transferability \cite{papernot2017practical,liu2017delving} across different models, \ie the adversarial examples generated on one model can still fool other models, which falls into the black-box attacks. 
% In this work, we mainly focus on the transfer-based attacks, and provide a brief introduction on related methods in this section.

% Szegedy \etal \cite{szegedy2014intriguing} first find that CNNs are vulnerable to aversarial examples, and propose a box-constrained L-BFGS method to craft adversarial examples. 
Existing white-box adversarial attacks \cite{szegedy2014intriguing,goodfellow2015FGSM,kurakin2017IFGSM,carlini2017towards,madry2018pgd} usually optimize the perturbation based on the gradient and exhibit good attack performance but low transferability. To boost the transferability, several gradient-based adversarial attacks have been proposed. Dong \etal \cite{dong2018boosting} propose to integrate momentum into iterative gradient-based attack. Lin \etal \cite{lin2020nesterov} propose to adopt Nestorve's accelerated gradient to further enhance the transferability. Liu \etal \cite{liu2017delving} have shown that ensemble-model attack, which attacks multiple models simultaneously can improve the transferability. 

Besides, recent works find that input transformations can further enhance the tranferability of adversarial attacks. Diverse Input Method (DIM) \cite{xie2019improving} creates diverse input patterns by applying random resizing and padding to the input at each iteration before feeding the image into the model for gradient calculation. Translation-Invariant Method (TIM)~\cite{dong2019evading} optimizes the perturbation over an ensemble of the translated images. To improve the efficiency, TIM convolves the gradient at the untranslated image with a pre-defined kernel, which needs one gradient calculation at each iteration. Scale-Invariant Method (SIM) \cite{lin2020nesterov} optimizes the adversarial perturbation over $m$ scale copies of the input to achieve higher transferability. 
\begin{figure*}
    \centering
    \includegraphics[width=\linewidth]{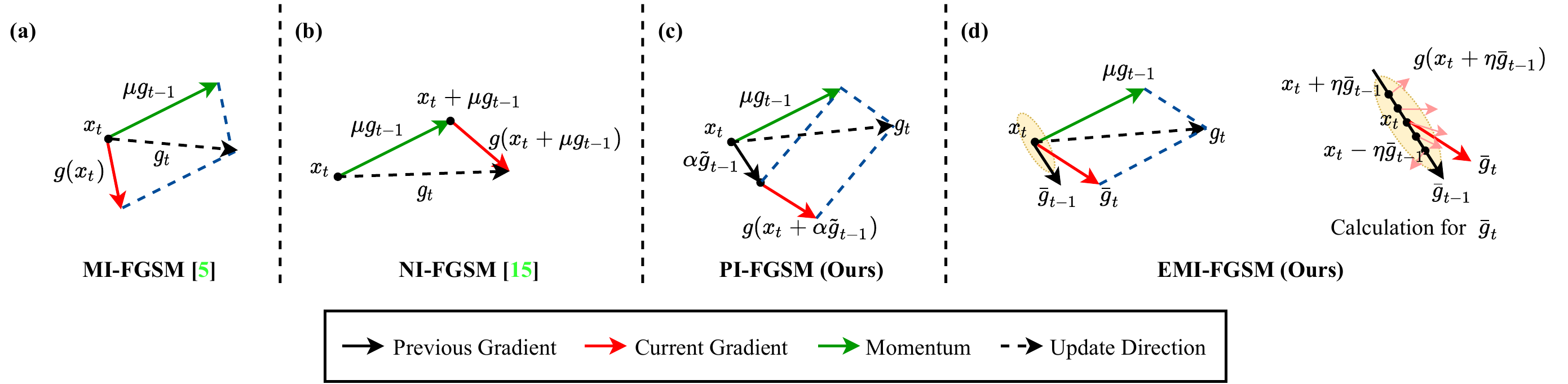}
    % \vspace{-0.5em}
    \caption{Illustration of the gradient update at the $t$-th iteration for various momentum based attack, where $g(x)$ denotes the gradient of $x$. (a) MI-FGSM~\cite{dong2018boosting} accumulates the gradient of $x_t$ for update. (b) NI-FGSM~\cite{lin2020nesterov} accumulates the gradient of $x_t +\mu g_{t-1}$ for update. (c) PI-FGSM  accumulates the gradient of $x_t + \alpha \tilde{g}_{t-1}$ for update, where $\tilde{g}_{t-1}$ is the gradient of the previous iteration. (d) EMI-FGSM accumulates the average gradient of the sampled data points in the direction of $\bar{g}_{t-1}$ for update, where $\bar{g}_{t-1}$ is the average gradient of the previous iteration.}
    \label{fig:gradient_update}
\end{figure*}
Both the ensemble-model attack and input transformation based attacks can be combined with gradient-based methods to further improve the transferability. Our method is a new variation of the gradient-based attack that exhibits higher transferability and can be integrated with the ensemble-model attack and input transformation based methods to achieve higher transferability.

\subsection{Adversarial Defenses}
As the counterpart of adversarial attacks, numerous works have been proposed to strengthen the robustness of deep learning models. 
Adversarial training \cite{goodfellow2015FGSM}, one of the most promising defense methods \cite{athalye2018obfuscated,li2019nattack}, which adopts adversarial examples for the training, has been extensively investigated by researchers~\cite{madry2018pgd, tramer2018ensemble, zhang2019theoretically, song2019improving, wang2020improving}. Among different adversarial training variations, ensemble adversarial training \cite{tramer2018ensemble}, which adopts the adversarial examples generated on ensemble models, has been demonstrated to be robust against transfer-based adversarial attacks. 
Besides, several denoising or input preprocessing methods have also been proposed to mitigate adversarial effects. Liao \etal \cite{liao2018defense} propose a high-level representation guided denoiser (HGD) for purifying the adversarial examples. Xie \etal \cite{xie2018mitigating} apply random resizing and padding (R\&P) to the input and feed the transformed image to the model to mitigate the adversarial effects. Feature distillation (FD) \cite{liu2019FD} is a JPEG-based defensive compression framework that can effectively rectify the adversarial examples without affecting the classification accuracy on benign images. Xu \etal \cite{xu2018BitReduction} propose to use bit depth reduction (Bit-Red) of each pixel for feature squeezing to detect adversarial examples. 
Moreover, some works focus on certified robustness. Cohen \etal \cite{cohen2019certified} propose to use randomized smoothing (RS) to train state-of-the-art certifiably ImageNet classifiers. Salman \etal \cite{salman2019provably} employ adversarial training to substantially improve the certified robustness of randomized smoothing (ARS).

\section{Methodology}
% \section{The Enhanced Momentum based Method}
In this section, we first give an overview on the family of 
gradient-based adversarial attacks, to which our method belongs. Then we provide detailed descriptions of the proposed Pre-gradient guided momentum Iterative FGSM (PI-FGSM) and Enhanced Momentum I-FGSM (EMI-FGSM).

\subsection{Gradient-based Adversarial Attacks}
%In recent years, several gradient-based adversarial attack methods have been proposed. %Here we give a brief overview.

Gradient-based adversarial attacks are typical methods for adversarial attacks.

% \textbf{Fast Gradient Sign Method (FGSM)} \cite{goodfellow2015FGSM} generates adversarial examples based on the hypothesis of the linearity of models and perform a one-step update:
\textbf{Fast Gradient Sign Method (FGSM)} \cite{goodfellow2015FGSM} generates adversarial examples by a one-step update:
\begin{equation*}
    x^{adv} = x + \epsilon \cdot \text{sign}(\nabla_x J_f(x, y)),
\end{equation*}
where $\text{sign}(\cdot)$ is the sign function and $\nabla_x J_f$ denotes the gradient of the loss function \wrt $x$.

\textbf{Iterative Fast Gradient Sign Method (I-FGSM)} \cite{kurakin2017IFGSM} extends FGSM by iteratively applying the gradient updates:
\begin{equation*}
    x_{t+1}^{adv} = x_t^{adv} + \alpha \cdot \text{sign}(\nabla_{x_t^{adv}} J_f(x_t^{adv}, y)),
\end{equation*}
where $x_1^{adv}=x$, $\alpha=\epsilon/T$ is a small step size, and $T$ is the number of iterations.

\textbf{Momentum Iterative Fast Gradient Sign Method (MI-FGSM)} \cite{dong2018boosting} proposes to integrate the momentum \cite{polyak1964some} into the iterative attack to achieve higher transferability:
\begin{gather*}
    g_t = \mu \cdot g_{t-1} + \frac{\nabla_{x_t^{adv}} J_f(x_t^{adv}, y)}{\|\nabla_{x_t^{adv}} J_f(x_t^{adv}, y)\|_1}, \\
    x_{t+1}^{adv} = x_t^{adv} + \alpha \cdot \text{sign}(g_{t}),
\end{gather*}
where $g_{t-1}$ is the accumulated gradient till the ($t-1$)-th iteration with a decay factor $\mu$ and $g_0 = 0$.

\textbf{Nesterov Iterative Fast Gradient Sign Method (NI-FGSM)} \cite{lin2020nesterov} integrates Nesterov's accelerated gradient (NAG) \cite{Nesterov1983} into the iterative attack method to further improve the transferability of adversarial examples:
\begin{gather}
    \tilde{x}_{t}^{adv} = x_{t}^{adv} + \alpha \cdot \mu \cdot g_{t-1}, \nonumber\\
    % \tilde{x}_{t}^{adv} = x_{t}^{adv} + \alpha \cdot \mu \cdot g_{t-1}, \label{eq:NAG}\\
    g_t = \mu \cdot g_{t-1} + \frac{\nabla_{\tilde{x}_t^{adv}} J_f(\tilde{x}_t^{adv}, y)}{\|\nabla_{\tilde{x}_t^{adv}} J_f(\tilde{x}_t^{adv}, y)\|_1}, \nonumber\\
    x_{t+1}^{adv} = x_t^{adv} + \alpha \cdot \text{sign}(g_t). \nonumber
\end{gather}

\subsection{Pre-gradient Guided Momentum based Attack}
\label{sec:PI-FGSM}
As shown in Figure~\ref{fig:gradient_update}~(a), MI-FGSM \cite{dong2018boosting} accumulates the gradient of each iteration to stabilize the update direction and escape from poor local maxima, and achieves higher transferability than I-FGSM~\cite{kurakin2017IFGSM}. As depicted in Figure~\ref{fig:gradient_update} (b), NI-FGSM \textit{looks ahead} by accumulating the gradient after adding momentum to the current data point so as to converge faster and achieve higher transferability~\cite{lin2020nesterov}. 

% MI-FGSM 利用全局信息平滑梯度
% NI-FGSM 向前看用局部信息会不会更好？

The performance improvement of NI-FGSM over MI-FGSM is mainly due to the \textit{looking ahead} property of the Nesterov’s  accelerated  gradient. We observe that NI-FGSM adopts the accumulated momentum in MI-FGSM to \textit{look ahead}, which is designed to obtain more stable direction by considering the history gradient. This inspires us to study a new problem: \textit{Although the direction of accumulated momentum helps craft more transferable adversaries, is it the optimal direction for looking ahead?}

% Different from typical optimization process, however, the optimization for the adversary generation has restricted the optima in the neighborhood of the original input. It raises us a new concern: \textit{With the accumulated momentum of NAG, the evaluated data point might exceed the neighborhood of the original input and provide imprecise gradient information, which may harm the transferability of the adversaries.} 

%To address this concern, 
To explore the direction of \textit{looking ahead}, we propose a variation of NI-FGSM, called the Pre-gradient guided momentum Iterative FGSM (PI-FGSM), which \textit{looks ahead} by the gradient of the previous iteration. Specifically, as shown in Figure \ref{fig:gradient_update}~(c), PI-FGSM accumulates the gradient of data point obtained by adding the previous gradient to the current data point at each iteration.
% adopts the gradient of the previous iteration to replace the ``look ahead'' portion that use the accumulated momentum for NAG as in Eq. \ref{eq:NAG}. 
The update procedure can be summarized as:
\begin{gather}
    \tilde{x}_t^{adv} = x_t^{adv} + \alpha \cdot \tilde{g}_{t-1},\quad
    \tilde{g}_{t}=\nabla_{\tilde{x}_t^{adv}} J_f(\tilde{x}_t^{adv}, y), \nonumber \\
    g_{t} = \mu \cdot g_{t-1} + \frac{\tilde{g}_{t}}{\|\tilde{g}_{t}\|_1},\quad
    x_{t+1}^{adv} = x_t^{adv} + \alpha \cdot \text{sign}(g_{t}), \nonumber
\end{gather}
where $\tilde{g}_{t-1}$ denotes the gradient of the previous iteration. Instead of considering all the history gradient as in NI-FGSM, PI-FGSM \textit{looks ahead} guided by the local gradient information and achieves better attack performance, as demonstrated in Sec.~\ref{sec:comp}.

\subsection{Enhanced Momentum based Attack}
\label{sec:EMI-FGSM}
We continue to investigate the family of momentum based attacks and observe that at each iteration, MI-FGSM~\cite{dong2018boosting}, NI-FGSM~\cite{lin2020nesterov} and our PI-FGSM accumulate the gradient of different data points, and they all exhibit higher transferability than I-FGSM~\cite{kurakin2017IFGSM} that only adopts the gradient of the current data point for update. 
This indicates that the accumulation of gradient is helpful for crafting highly transferable adversaries. 
%Since all the above methods only accumulate gradient of data points in the previous iterations, 
Since the accumulation of gradient of these methods are on different data points, 
this inspires us another question: \textit{At each iteration, could we further improve the attack transferability by accumulating the gradients of multiple data points around the current data point for the iterative gradient-based attacks?}
% On the other hand, though MI-FGSM and NI-FGSM accumulates the gradient of various data points in the previous iterations, they both achieve higher transferability. This inspires us another question: %concern: 
% \textit{At each iteration, could we further improve the attack transferability by accumulating the gradients of multiple data points around the current data point for the iterative gradient-based attacks?}
% To address this question, we further add sampled data points in the direction of the gradient at the previous iteration and get the Enhanced MI-FGSM (EMI-FGSM), as depicted in Sec.~\ref{sec:EMI-FGSM}.

To address this question, we enhance the momentum by not only memorizing all the past gradients during the iterative process, but also accumulating the gradients of multiple sampled examples 
%around the current data point. 
in the vicinity of the current data point. 
% To help sample more useful data points for the gradient calculation, we sample data points in the gradient direction of the previous iteration. 
Considering the performance improvement of PI-FGSM, to help sample more useful data points for the gradient calculation, we sample multiple data points along the direction used in PI-FGSM, \ie the gradient direction of the previous iteration. Specifically, as shown in Figure~\ref{fig:gradient_update} (d), we calculate the gradient of the $t$-th iteration as follows:
%  instead of randomly sampling examples in the neighborhood of the input as in RMI-FGSM,
\begin{gather}
    \bar{x}_t^{adv}[i] = x_t^{adv} + c_i \cdot \bar{g}_{t-1} \label{eq:sample_data} \\
    \bar{g}_{t} = \frac{1}{N}\sum_{i=1}^N \nabla_{\bar{x}_t^{adv}[i]} J_f(\bar{x}_t^{adv}[i], y),
    \label{eq:gradient}
\end{gather}
where $N$ is the sampling number, $\bar{g}_{t-1}$ is the gradient calculated at the previous iteration and $c_i$ is the $i$-th coefficient sampled in interval $[-\eta, \eta]$. We denote such accumulated gradient as the enhanced momentum.

Note that the proposed enhanced momentum is generally applicable to any iterative gradient-based attacks, such as I-FGSM \cite{kurakin2017IFGSM}, PGD \cite{madry2018pgd}, and the ensemble-model attack~\cite{liu2017delving}. Here we  
incorporate the enhanced momentum into I-FGSM, denoted as Enhanced Momentum I-FGSM (EMI-FGSM), to craft highly transferable adversarial examples. The update procedure can be summarized as:
\begin{gather}
    g_{t} = \mu \cdot g_{t-1} + \frac{\bar{g}_{t}}{\|\bar{g}_{t}\|_1}, \label{eq:momentum_update} \\
    x_{t+1}^{adv} = x_t^{adv} + \alpha \cdot \text{sign}(g_{t}), \label{eq:adv_update}
\end{gather}
where $\bar{g}_{t}$ is the gradient calculated by Eq.~\eqref{eq:gradient}. The algorithm of EMI-FGSM is summarized in Algorithm~\ref{alg:EMI-FGSM}. 
%Except for sampling the data points in the direction of PI-FGSM, 
For other possible sampling methods, 
we have also tried to sample the data points in the direction of NI-FGSM or do random sampling, 
but the sampling in the direction of PI-FGSM exhibits better results, as discussed in Sec.~\ref{sec:discussion}.

%\subsection{The Relationship of Various Methods}
%As shown in Table~\ref{tab:taxonomy}, the existing transfer-based adversarial attacks can roughly fall into two categories: 1) Optimization-based methods adopt various gradient calculation methods (\eg momentum, NAG, enhanced momentum \etc) to enhance the transferability. 2) Transformation-based methods employ various input transformations (\eg random resizing and padding, translation, scale \etc) before feeding the image into the deep models for gradient calculation. The proposed EMI-FGSM falls into optimization-based methods because it utilizes a better optimization algorithm to generate adversarial examples. Optimization-based methods can be combined with transformation-based methods to achieve higher transferability as in \cite{xie2019improving, dong2019evading, lin2020nesterov}. Both optimization-base methods and transformation-based methods can attack the ensemble of models to further boost the adversarial attacks.

%  MPI-FGSM can also be combined with transformation-based attack methods (\eg DIM, TIM and SIM), to further boost the adversarial attack.

%\input{tables/taxonomy_of_gradient_based_adversarial_attack_methods}

\setlength{\textfloatsep}{1.em}
\begin{algorithm}[tb]
    \algnewcommand\algorithmicinput{\textbf{Input:}}
    \algnewcommand\Input{\item[\algorithmicinput]}
    \algnewcommand\algorithmicoutput{\textbf{Output:}}
    \algnewcommand\Output{\item[\algorithmicoutput]}
    \caption{EMI-FGSM.}
    \label{alg:EMI-FGSM}
	\begin{algorithmic}[1]
		\Input A classifier $f$ and loss function $J_f$. A benign example $x$ and its ground-truth label $y$.
		\Input The maximum perturbation $\epsilon$, number of iteration $T$ and decay factor $\mu$. The bound $\eta$ for the sampling interval and sampling number $N$.
        \Output An adversarial example $x^{adv} \in \mathcal{B}_{\epsilon}(x)$.
        % where $\|x^{adv}-x\|_\infty \leq \epsilon$.
		\State $\alpha = \epsilon/T$; $g_0 = 0$; $\bar{g}_0 = 0$; $x_1^{adv}=x$.
		\For{$t = 1 \rightarrow T$}:
		    \State Sample $N$ coefficients $c_i \in [-\eta, \eta]$ for Eq.~\eqref{eq:sample_data}.
		    \State Calculate the average gradient $\bar{g}_{t}$ of the $N$ sampled data points in the neighborhood of $x_t^{adv}$ by Eq.~\eqref{eq:gradient}.% with $c_i$.
		    \State Update the enhanced momentum $g_{t}$ by Eq.~\eqref{eq:momentum_update}.
		    \State Update $x_{t+1}^{adv}$ by Eq.~\eqref{eq:adv_update}.
		  %  \State Update $g_{t}$ by accumulating the momentum by Eq.~\eqref{eq:momentum_update}.
		  %  \State Update $x_{t+1}^{adv}$ by the direction of momentum $g_{t}$ by Eq.~\eqref{eq:adv_update}.
		\EndFor
        \State \Return $x^{adv}=x_{T+1}^{adv}$.
	\end{algorithmic} 
\end{algorithm} 
% \vspace{-1em}
\begin{table*}[tb]
\begin{center}
% \resizebox{\textwidth}{!}{
\scalebox{0.90}{
\begin{tabular}{l|cccccccc} 
\hline
Attack & Inc-v3* & Inc-v4 & IncRes-v2 & Res-101 & Inc-v3${\rm _{ens3}}$ & Inc-v3${\rm _{ens4}}$ & IncRes-v2${\rm _{ens}}$ \\ 
\hline \hline
FGSM & ~~67.3 & 25.7 & 26.0 & 24.5 & 10.2 & 10.4 & ~~4.5 \\     
I-FGSM & \textbf{100.0} & 20.3 & 18.5 & 16.1 & ~~4.6 & ~~5.2 & ~~2.5 \\
MI-FGSM & \textbf{100.0} & 44.5 & 42.0 & 36.3 & 13.4 & 13.7 & ~~6.5 \\
NI-FGSM & \textbf{100.0} & 51.9 & 50.4 & 41.0 & 13.4 & 13.2 & ~~5.7 \\
% RI-FGSM & ~~99.7 & 44.0 & 41.5 & 35.3 & 13.5 & 13.1 & ~~6.1 & 36.2 \\
\hline
PI-FGSM (\textbf{Ours}) & \textbf{100.0} & 60.2 & 59.1 & 49.0 & 14.9 & 14.6 & ~~6.5 \\
% MNI-FGSM & \textbf{100.0} & 50.9 & 49.0 & 41.2 & 14.2 & 14.0 & ~~7.2 & 39.5 \\
%MRI-FGSM & ~~99.1 & 61.1 & 59.3 & 52.0 & \textbf{23.1} & \textbf{23.0} & \textbf{11.5} & 47.0 \\
EMI-FGSM (\textbf{Ours}) & \textbf{100.0} & \textbf{72.7} & \textbf{69.9} & \textbf{59.5} & \textbf{20.3} & \textbf{19.9} & \textbf{10.9} \\
\hline
\end{tabular}
}
\vspace{-0.3em}
\caption{\textbf{Attack success rates (\%) of adversarial attacks against the seven baseline models under single-model setting.} The adversarial examples are crafted on Inc-v3. \textbf{* indicates the white-box model being attacked.}}
% \han{[han: I suggest mark also PI-FGSM as your methods at some place, either within the table or at the caption. Without carefully reading the main text, I may easily miss the fact that PI is also your contribution.]}
\label{tab:comparison_inc_v3}
\end{center}
\vspace{-1.4em}
\begin{center}
\scalebox{0.95}{
\begin{tabular}{l|cccccccc} 
\hline
Attack & Inc-v3* & Inc-v4* & IncRes-v2* & Res-101* & Inc-v3${\rm _{ens3}}$ & Inc-v3${\rm _{ens4}}$ & IncRes-v2${\rm _{ens}}$ \\ 
\hline \hline
FGSM &  ~~64.8 & 49.3 & 43.9 & ~~68.8 & 15.8 & 15.1 & ~~8.9 \\
I-FGSM & ~~99.9 & 98.6 & 95.6 & ~~99.8 & 19.1 & 16.8 & 10.4 \\
MI-FGSM & ~~99.9 & 98.7 & 95.0 & ~~99.9 & 39.7 & 35.5 & 23.8 \\
NI-FGSM & \textbf{100.0} & \textbf{99.8} & 99.2 & ~~99.9 & 41.2 & 34.9 & 22.9 \\
% RI-FGSM & 99.9 & 98.6 & 95.3 & 99.9 & 39.7 & 36.0 & 25.2 & 70.7 \\
\hline
PI-FGSM (\textbf{Ours}) & \textbf{100.0} & 99.2 & 98.5 & ~~99.9 & 52.5 & 45.3 & 29.6 \\
% ENI-FGSM & \textbf{100.0} & \textbf{99.9} & \textbf{99.9} & \textbf{100.0} & 47.4 & 42.4 & 27.3 & 73.8 \\
% ERI-FGSM & 99.8 & 98.9 & 96.7 & 99.8 & 64.6 & 59.4 & \textbf{46.6} & 80.8 \\
EMI-FGSM (\textbf{Ours}) & \textbf{100.0} & \textbf{99.8} & \textbf{99.8} & \textbf{100.0} & \textbf{69.0} & \textbf{62.0} & \textbf{43.0} \\
\hline
\end{tabular}
}
\vspace{-0.3em}
\caption{\textbf{Attack success rates (\%) of adversarial attacks against the seven baseline models under ensemble-model setting.} The adversarial examples are crafted on ensemble models including Inc-v3, Inc-v4, IncRes-v2 and Res-101.}
\label{tab:comparison_ens}
\end{center}
\vspace{-1.6em}
\end{table*}
\section{Experiments}
%In this section, we present the experimental results to validate the efficacy of the proposed method. We first specify the experimental setup in Sec.~\ref{sec:exp_setup}. Then we provide results of various optimization-based attacks on four normally trained models in Sec.~\ref{sec:comp}. We further integrate our method into the transformation-based attacks and ensemble-model attack in Sec.~\ref{sec:exp_trans} and attack seven advanced defense models in Sec.~\ref{sec:exp_defense}. Finally, we provide ablation study for the sampling method and hyper-parameters in Sec.~\ref{sec:exp_ablation} and further discussion on other possible variant methods in Sec.~\ref{sec:discussion}. 

%We have conducted abundant experiments to verify the effectiveness of the proposed method. 
In this section, we provide the experimental setup, report comparisons of gradient-based attacks on four normally trained models and comparisons when integrated with transformation-based attacks and ensemble-model attack, as well as results of attacking seven advanced defense models. In the end, we further provide ablation studies for the sampling method and hyper-parameters as well as discussions on other possible variant methods. 
%validate the efficacy of the proposed method. 
%In this section, we first provide experimental setup, and the results of gradient-based attacks on four normally trained models. Then we integrate our method into the transformation-based attacks and ensemble-model attack, and attack seven advanced defense models. Finally, we provide ablation studies for the sampling method and hyper-parameters and further discussion on other possible variant methods. 

\subsection{Experimental Setup}
\label{sec:exp_setup}
\textbf{Dataset.} Similar to \cite{xie2019improving, lin2020nesterov}, we randomly choose 1,000 images from the ILSVRC 2012 validation set \cite{russakovsky2015imagenet}. All these images are resized to $299\times299\times3$ beforehand.

% \textbf{Baselines.} We compare our proposed method with two momentum-based iterative methods, namely MI-FGSM \cite{dong2018boosting} and NI-FGSM \cite{lin2020nesterov}, which have shown great transferbaility. Moreover, to demonstrate the efficacy of our method, we also adopt four similar variations of MI-FGSM, using the gradient of the input with random noise denoted as RI-FGSM, using the gradient of the input applied by the gradient of the previous iteration, denoted as PI-FGSM, and extending NI-FGSM and RI-FGSM to accumulate the average gradient as EMI-FGSM, denoted as NMI-FGSM and RMI-FGSM respectively.
% % using the average gradient of images randomly sampled from the neighborhood of the input, denoted RMI-FGSM respectively. 
% The details of the four variations are provided in Appendix~\ref{sec:variation}. We also integrate the proposed method into the ensemble attack \cite{liu2017delving, dong2018boosting} and input transformations \cite{xie2019improving, dong2019evading, lin2020nesterov} as in \cite{lin2020nesterov}.

\textbf{Baselines.} We compare our method with four gradient-based attack methods including FGSM \cite{goodfellow2015FGSM}, I-FGSM \cite{kurakin2017IFGSM}, MI-FGSM \cite{dong2018boosting} and NI-FGSM \cite{lin2020nesterov}. We also integrate our method into the ensemble-model attack \cite{liu2017delving, dong2018boosting} and input transformation based methods \cite{xie2019improving, dong2019evading, lin2020nesterov}, to show the performance improvement of our method over these baselines.
% \textbf{Baselines.} We compare the proposed method with two momentum-based iterative methods, \ie MI-FGSM \cite{dong2018boosting} and NI-FGSM \cite{lin2020nesterov}, which exhibit the state-of-the-art transferbaility. We also integrate the proposed method into the ensemble-model attack \cite{liu2017delving, dong2018boosting} and input transformation methods \cite{xie2019improving, dong2019evading, lin2020nesterov} to further promote the transferability.

\textbf{Models.} Four normally trained models, \ie Inception-v3 (Inc-v3) \cite{szegedy2016inceptionv3}, Inception-v4 (Inc-v4), Inception-Resnet-v2 (IncRes-v2) \cite{szegedy2017inception}, Resnet-v2-101 (Res-101) \cite{he2016resnet}, as well as three ensemble adversarially trained models, \ie ens3-adv-Inception-v3 (Inc-v3$_{ens3}$), ens4-Inception-v3 (Inc-v3$_{ens4}$), ens-adv-Inception-ResNet-v2 (IncRes-v2$_{ens}$) \cite{tramer2018ensemble}, are considered.  
Without ambiguity, we simply call the three ensemble adversarially trained models as \textit{adversarially trained models}. 
Moreover, to show the efficacy of our methods, we also incorporate seven advanced defense methods, including the top-3 submission in the NIPS 2017 defense competition, \ie high-level representation guided denoiser (HGD, rank-1) \cite{liao2018defense}, input transformation through random resizing and padding (R\&P, rank-2) \cite{xie2018mitigating}, NIPS-r3 (rank-3) \footnote{\url{https://github.com/anlthms/nips-2017/tree/master/mmd}}, randomized smoothing (RS) \cite{cohen2019certified} and adversarially randomized smoothing (ARS) \cite{salman2019provably} for certified defense, feature distillation (FD) \cite{liu2019FD} and bit depth reduction (Bit-Red) \cite{xu2018BitReduction}.

\textbf{Attack Settings.} We follow the settings in \cite{dong2018boosting} with the maximum perturbation of $\epsilon = 16/255$, pixel values normalized into $[0,1]$ and the number of iteration $T=10$. For the momentum term, we set the decay factor $\mu = 1$ as in~\cite{dong2018boosting, lin2020nesterov}. For DIM, we set the transformation probability to $0.5$ and the input $x$ is first randomly resized to an $r \times r \times 3$ image with $r \in [299, 330)$, and then padded to size $330 \times 330 \times 3$ as in \cite{xie2019improving}. For TIM, we adopt Gaussian kernel with size $7 \times 7$ as in \cite{dong2019evading}. For SIM, the number of scale copy is set to $m=5$ as in \cite{lin2020nesterov}. For EMI-FGSM, we set the number of examples $N$ to 11, set the sampling interval bound $\eta=7$, and adopt the linear sampling.

% \textbf{Attack Setting.} We follow the setting in \cite{dong2018boosting} with the maximum perturbation $\epsilon = 16/255$ with pixel values normalized into $[0,1]$ and number of iteration $T=10$. For the momentum term, we set the decay factor $\mu = 1$ as in \cite{dong2018boosting, lin2020nesterov}. For DIM, we set the transformation probability as $0.5$ and the input $x$ is first randomly resized to a $rnd \times rnd \times 3$ image with $rnd \in [299, 330)$, and then padded to the size $330 \times 330 \times 3$ as in \cite{xie2019improving}. For TIM, we adopt Gaussian kernel with the size of $7 \times 7$. For SIM, the number of scale copies is set to $m=5$. The number of examples $N$ for NMI-FGSM, RMI-FGSM and EMI-FGSM is set to 11. For NMI-FGSM and EMI-FGSM, we set the sampling interval $\eta=7$ and adopt linear sampling.

\subsection{Comparison with Gradient-based Attacks}
\label{sec:comp}
We first craft adversaries by various gradient-based attacks under single-model setting and ensemble-model setting respectively, and report the attack success rates, which are the misclassification rates of the corresponding models using adversarial examples as the inputs. 

\textbf{Single-model Setting.} The results for adversarial examples crafted on Inc-v3 are depicted in Table~\ref{tab:comparison_inc_v3} and the results on other three normally trained models are summarized in Appendix.
We can see that except for FGSM, all the other attacks achieve 100\% attack success rates under white-box setting. For black-box attacks, I-FGSM achieve the transferability even lower than FGSM. Compared with MI-FGSM and NI-FGSM, the transferability of the proposed PI-FGSM is much higher (8-9\%) on normally trained models, and is considerably higher (0.8-1.5\%) on adversarially trained models. 
With the enhanced momentum, EMI-FGSM exhibits much higher transferability on both normally trained models (10.5-12.5\% higher than PI-FGSM) and adversarially trained models (4.4-5.4\% higher than PI-FGSM), and outperforms the powerful baseline NI-FGSM with a clear margin of 11.1\% on average.

\textbf{Ensemble-model Setting.} As in \cite{dong2018boosting}, we implement the attacks under ensemble-model setting by fusing the logit outputs of four normally trained models, \ie Inc-v3, Inc-v3, IncRes-v2 and Res-101, with equal ensemble weights. As shown in Table~\ref{tab:comparison_ens}, the proposed PI-FGSM exhibits better attack success rates than I-FGSM and MI-FGSM under white-box setting and achieves much higher transferability on three adversarially trained models. This validates our first concern that due to considering too much history gradient, the accumulated momentum adopted by NI-FGSM provide imprecise direction compared with PI-FGSM, which is not optimal for looking ahead.
% the evaluated data point of NI-FGSM might exceed the neighboorhood of the original input and harm the transferability. 
Moreover, EMI-FGSM achieve the best results under both white-box and black-box setting and outperforms the powerful baseline NI-FGSM by a large margin of more than 20\%, which demonstrates the high effectiveness of the proposed enhanced momentum.

%We also illustrate two randomly sampled adversarial images generated on Inc-v3 model by MI-FGSM, NI-FGSM and EMI-FGSM in Figure~\ref{fig:adv_images}. It can be seen that the perturbations crafted by EMI-FGSM are smoother than that of MI-FGSM and NI-FGSM, which might be the reason why EMI-FGSM achieves such great transferability.

% by using EMI-FGSM, in which the enhanced momentum accumulate the gradients of the
% sampled examples, the adversarial perturbations are smoother than those generated by MI-FGSM and NI-FGSM. The smooth effect also exists in other enhance momentum based attacks.
\begin{table*}[tb]
\begin{center}
% \resizebox{\textwidth}{!}{
 \scalebox{0.9}{
\begin{tabular}{l|lccccccc}
\hline
Attack & Inc-v3* & Inc-v4 & IncRes-v2 & Res-101 & Inc-v3$_{ens3}$ & Inc-v3$_{ens4}$ & IncRes-v2$_{ens}$\\
\hline\hline
DIM & ~~99.0 & 64.6 & 60.9 & 52.1 & 18.3 & 17.7 & ~~9.5 \\
EMI-DIM (\textbf{Ours}) & \textbf{~~99.1} & \textbf{83.5} & \textbf{78.0} & \textbf{70.6} & \textbf{27.8} & \textbf{26.0} & \textbf{13.4} \\
\hline
TIM & \textbf{100.0} & 47.0 & 44.5 & 40.5 & 24.3 & 22.0 & 13.2 \\
EMI-TIM (\textbf{Ours}) & \textbf{100.0} & \textbf{79.4} & \textbf{76.3} & \textbf{67.2} & \textbf{44.3} & \textbf{40.8} & \textbf{26.2} \\
\hline
SIM & \textbf{100.0} & 70.3 & 68.0 & 62.4 & 32.4 & 30.8 & 17.2 \\
EMI-SIM (\textbf{Ours}) & \textbf{100.0} & \textbf{91.9} & \textbf{90.0} & \textbf{85.4} & \textbf{45.2} & \textbf{41.8} & \textbf{23.8} \\
\hline
DTS & ~~98.9 & 83.1 & 80.7 & 75.8 & 65.2 & 62.7 & 46.0 \\
EMI-DTS (\textbf{Ours}) & \textbf{~~99.6} & \textbf{94.1} & \textbf{92.6} & \textbf{89.4} & \textbf{78.9} & \textbf{75.3} & \textbf{60.4} \\
\hline
\end{tabular}
 }
\vspace{-0.4em}
\caption{\textbf{Attack success rates (\%) of adversarial attacks against the seven baseline models under single-model setting.} The adversarial examples are crafted on Inc-v3.}
\label{tab:singleModel}
\end{center}
\vspace{-1.4em}
\end{table*}
\begin{table*}[tb]
\begin{center}
% \resizebox{\textwidth}{!}{
\scalebox{0.90}{
\begin{tabular}{l|ccccccc}
\hline
Attack & Inc-v3* & Inc-v4* & IncRes-v2* & Res-101* & Inc-v3$_{ens3}$ & Inc-v3$_{ens4}$ & IncRes-v2$_{ens}$\\
\hline\hline
DIM & ~~99.4 & ~~97.4 & ~~94.7 & \textbf{~~99.8} & 56.3 & 50.7 & 36.4 \\
EMI-DIM (\textbf{Ours}) & \textbf{~~99.9} & \textbf{~~99.6} & \textbf{~~99.7} & ~~99.7 & \textbf{77.0} & \textbf{70.1} & \textbf{50.3} \\
\hline
TIM & ~~99.8 & ~~98.0 & ~~95.0 & ~~99.9 & 61.3 & 56.7 & 47.8 \\
EMI-TIM (\textbf{Ours}) & \textbf{100.0} & \textbf{100.0} & \textbf{~~99.7} & \textbf{100.0} & \textbf{89.0} & \textbf{83.9} & \textbf{78.2} \\
\hline
SIM & ~~99.9 & ~~99.3 & ~~98.5 & \textbf{100.0} & 78.5 & 74.4 & 60.4 \\
EMI-SIM (\textbf{Ours}) & \textbf{100.0} & \textbf{100.0} & \textbf{100.0} & \textbf{100.0} & \textbf{90.1} & \textbf{87.3} & \textbf{74.2} \\
\hline
DTS & ~~99.6 & ~~98.9 & ~~97.9 & ~~99.7 & 92.1 & 90.2 & 86.6 \\
EMI-DTS (\textbf{Ours}) & \textbf{100.0} & \textbf{~~99.9} & \textbf{100.0} & \textbf{100.0} & \textbf{97.4} & \textbf{96.1} & \textbf{94.1} \\
\hline
\end{tabular}
}
\vspace{-0.4em}
\caption{\textbf{Attack success rates (\%) of adversarial attacks against the seven baseline models under ensemble-model setting.} The adversarial examples are crafted on ensemble models including Inc-v3, Inc-v4, IncRes-v2 and Res-101.}
\label{tab:ensembleModel}
\end{center}
\vspace{-1.7em}
\end{table*}
\subsection{Integrated with Transformation-based Attacks}%Extension to
\label{sec:exp_trans}  %Incorporation
%To further show the efficacy of our enhanced momentum based method, we 
We further incorporate EMI-FSGM with various input transformations, \ie DIM, TIM, SIM, and the combination of the three input transformations, denoted as 
%DI-TI-SIM (DTS for abbreviation),
DTS for abbreviation, under single-model setting and ensemble-model setting respectively. To ensure fairness, all the transformations are integrated into MI-FGSM as baselines \cite{xie2019improving, dong2019evading}. 

\textbf{Single-model Setting.} The results for adversaries generated on Inc-v3 are summarized in Tabel~\ref{tab:singleModel}. We can observe that EMI can significantly boost the transferability on each of the transformation-based attack methods. In general, the
EMI based attacks consistently outperform the baseline attacks by $3.9\% \sim 32.4\%$.
Even for white-box setting, EMI further promotes the attack success rates of the baseline attacks. For instance, EMI-DTS outperforms DTS by $0.7\%$ against Inc-v3.
The results for adversarial examples crafted on other three normally-trained models are consistent with that generated on Inc-v3, that are summarized in Appendix.

% \textbf{Single-model attacks.} The attack success rates under single model setting are summarized in Table~\ref{tab:singleModel}. As can be observed, our EMI method can significantly boost the transferability on each of the transformation-based attack methods.
% Under black-box setting, EMI significantly promotes the attack success rates of the baseline attacks by a clear margin.
% Even for white-box setting, EMI further promotes the attack success rates of the baseline attacks, and yields an average attack success rate in the range of 98.5\%-100\%.\HK{Where is 98.5\% in this table?}
% We take the attacks on Inc-v3 model as an example.\HK{This is exactly in this table! Rewrite the logic of this parapraph.} 
% EMI outperforms the baselines by at least $17\%$ for DIM, $27\%$ for TIM, $21\%$ for SIM, and $11\%$ for DTS against three normally trained models. Also, EMI outperforms the baselines by at least $4\%$ for DIM, $13\%$ for TIM, $6\%$ for SIM, and $12\%$ for DTS against the three adversarially trained models. The results for adversaries generated on other three normally-trained models are summarized in Appendix (Table \ref{tab:comparison_three_models}).
%\HK{add Table ?}.
%The results validate the effectiveness of the propose method.

\textbf{Ensemble-model Setting.} As in Sec.~\ref{sec:comp}, we also evaluate the attacks under ensemble-model setting and the results are summarized in Table~\ref{tab:ensembleModel}. EMI based method remarkably improves the attack success rates across all experiments over the baseline attacks. 
In particular, the final combination of EMI-DTS has achieved the attack success rates of over 94.1\% for black-box attacks against the three adversarially trained models.
Such intriguing results convincingly demonstrate the %high efficacy
success on the combination of EMI-FGSM, input transformations and ensemble-model attack for improving the attack transferability.

\begin{figure*}
    \centering
    \begin{subfigure}{.24\textwidth}
          \centering 
          \includegraphics[width=\linewidth]{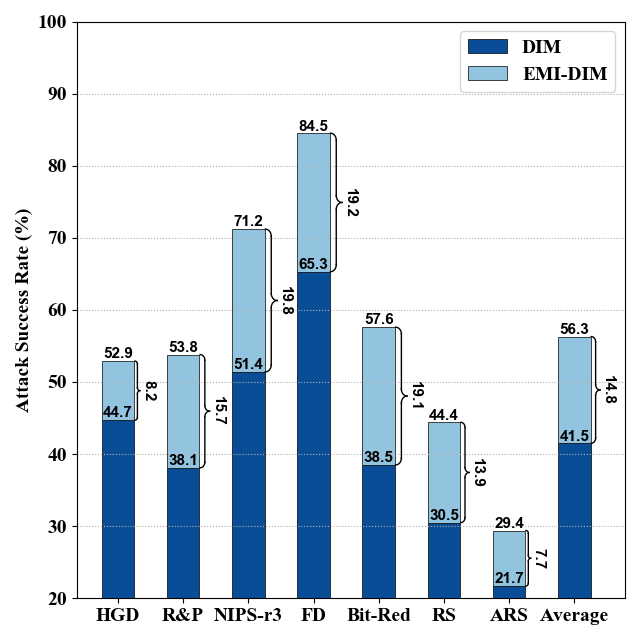}
          
          \caption{DIM \vs EMI-DIM}
        \label{fig:defense:DIM}
    \end{subfigure}
    % \hspace{-1.5em}
    \begin{subfigure}{.24\textwidth} 
      \centering 
      \includegraphics[width=\linewidth]{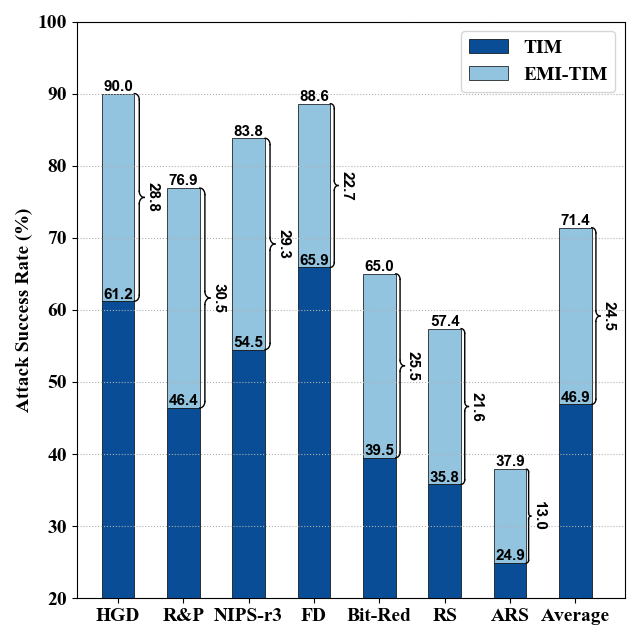}
      
      \caption{TIM \vs EMI-TIM}
      \label{fig:defense:TIM}
    \end{subfigure}
    % \hspace{-1.5em}
    \begin{subfigure}{.24\textwidth}
      \centering 
      \includegraphics[width=\linewidth]{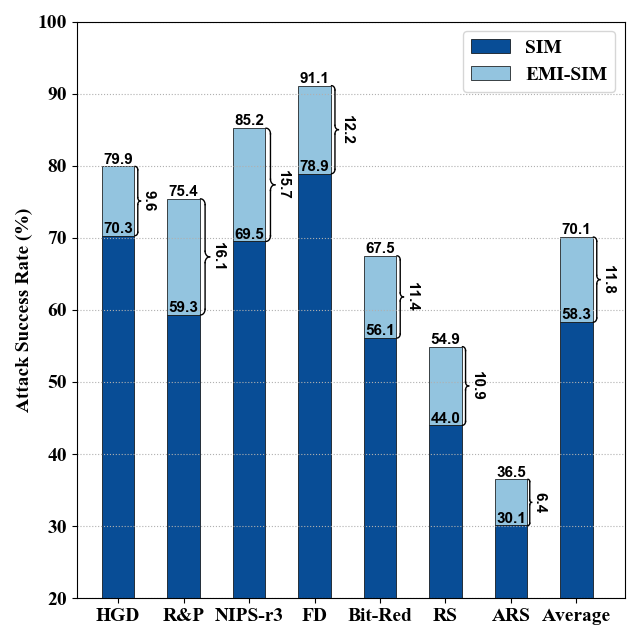}
      
      \caption{SIM \vs EMI-SIM}
      \label{fig:defense:SIM}
    \end{subfigure}
    % \hspace{-1.5em}
    \begin{subfigure}{.24\textwidth}
      \centering 
      \includegraphics[width=\linewidth]{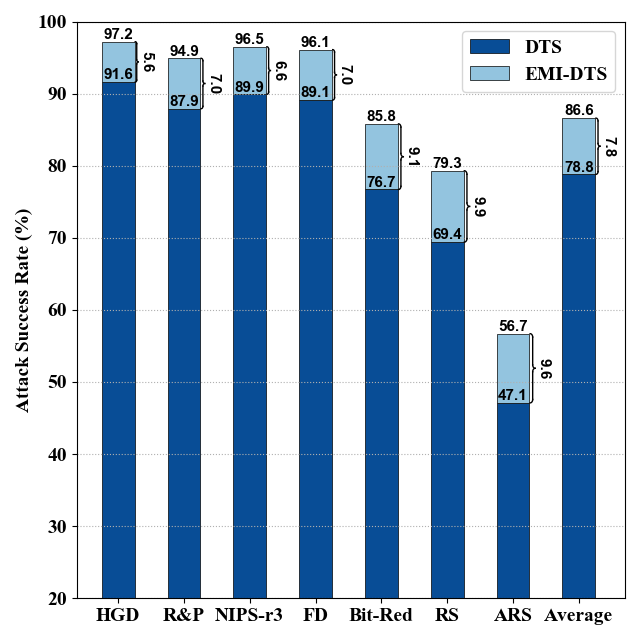}
      
      \caption{DTS \vs EMI-DTS}
      \label{fig:defense:DTS}
    \end{subfigure}
    \vspace{-0.5em}
    \caption{\textbf{Attack success rates (\%) of adversarial attacks against seven advanced defense methods.} The adversarial examples are crafted on the ensemble models including Inc-v3, Inc-v4, IncRes-v2, and Res-101. (Zoom in for details.)}
    \label{fig:defense}
\end{figure*}
\subsection{Attacking Advanced Defense Models}
\label{sec:exp_defense}
With the significant improvement on the attack baselines, we further evaluate EMI-FGSM on seven advanced defense models with various input transformations under ensemble-model setting to demonstrate its high efficacy. All the adversaries are generated on the ensemble models as in Sec.~\ref{sec:exp_trans} and we test the advanced defenses with these adversaries.

The results of EMI-FGSM with three transformation-based attacks are illustrated in Figure~\ref{fig:defense:DIM}-\ref{fig:defense:SIM}. As can be observed, EMI-FGSM remarkably improves the transferability of the three transformation-based attacks on all these defense equipped models. On average, the performances are improved by 14.8\%, 24.5\% and 11.8\% respectively. Moreover, we also integrate the combination of the three input transformations into EMI-FGSM as in \cite{lin2020nesterov} to further improve the transferability. As shown in Figure~\ref{fig:defense:DTS}, EMI-DTS achieves an average attack success rate of 86.6\%, which boosts existing state-of-the-art methods by a clear margin of 7.8\%. Considering that the adversaries are crafted on the ensemble models without any defense mechanisms but with such high attack performance, the results identify the inefficiency of existing defenses and indicate that they are far from being deployed in real-world applications. 
% \HK{Suggest to add ``(See detailed numbers in Appendix, Table ?.)"}

% \HK{In experiments, you only list results, but did not provide some insights on why it is good, why the combination is good.}

\begin{table*}[t]
\begin{center}
% \resizebox{\textwidth}{!}{
\scalebox{0.90}{
\begin{tabular}{c|c|ccccccc} 
\hline
Attack & Sampling Method & Inc-v3* & Inc-v4 & IncRes-v2 & Res-101 & Inc-v3${\rm _{ens3}}$ & Inc-v3${\rm _{ens4}}$ & IncRes-v2${\rm _{ens}}$ \\ 
\hline \hline
\multirow{3}{*}{EMI-FGSM} & Linear & \textbf{100.0} & 74.4 & \textbf{71.9} & 60.5 & \textbf{21.2} & \textbf{19.5} & \textbf{~~9.9} \\
& Uniform & \textbf{100.0} & \textbf{74.7} & 71.5 & \textbf{61.2} & 18.9 & 18.8 & ~~8.8 \\
& Gaussian & \textbf{100.0} & 73.0 & 70.4 & 60.0 & 20.2 & 19.0 & \textbf{~~9.9} \\
\hline \hline
\multirow{3}{*}{EMI-DTS} & Linear & \textbf{~~99.6} & 94.5 & \textbf{92.8} & \textbf{90.2} & \textbf{78.8} & \textbf{76.0} & \textbf{60.3} \\
& Uniform & ~~99.5 & 92.9 & 91.9 & 87.9 & 77.1 & 71.5 & 57.0 \\
& Gaussian & ~~99.5 & \textbf{94.8} & 92.6 & 89.7 & 78.5 & 74.3 & 58.7  \\
\hline
\end{tabular}
}
\vspace{-0.3em}
\caption{\textbf{Attack success rates (\%) of EMI-FGSM and EMI-DTS against  the seven baseline models with various sampling methods.} The adversarial examples are crafted on Inc-v3.}
\label{tab:sampleMethod}
\end{center}
\vspace{-1em}
\end{table*}
\subsection{Ablation Study}
\label{sec:exp_ablation}
%To further gain insights
To gain more insights on the performance improvement by our enhanced momentum based methods, we conduct ablation studies to explore the impact of the sampling method and the hyper-parameters for the sampling interval $\eta$ and sampling number $N$, respectively. To simplify the analysis, we only consider the transferability of adversarial examples crafted on Inc-v3 model by vanilla EMI-FGSM and EMI-DTS. The default setting adopts linear sampling, and we set $\eta=7$ and $N=11$.

% \HK{I suggest we add a summary on the following three paragraphs and put the following in Appendix. You may say that our EMI methods are not very sensitive for some parameters, like the sampling distribution and sampling interval. }

\textbf{On sampling distribution.} We first report the results of EMI-FGSM and EMI-DTS with three types of sampling methods, \ie, linear sampling, uniform sampling and Gaussian sampling. 
Linear sampling samples $N$ linearly spaced data points in the interval. Uniform sampling and Gaussian sampling sample $N$ data points in the interval by uniform distribution and Gaussian distribution, respectively.
As shown in Table~\ref{tab:sampleMethod}, the three sampling methods achieve similar attack performance, showing much higher transferability than the attack baselines. In general, linear sampling exhibits slightly higher results, thus we adopt linear sampling in experiments.

\begin{figure*}[t]
\centering 
    \begin{minipage}[b]{.48\textwidth} 
        \begin{subfigure}{.48\textwidth}
          \centering 
          \includegraphics[width=\linewidth]{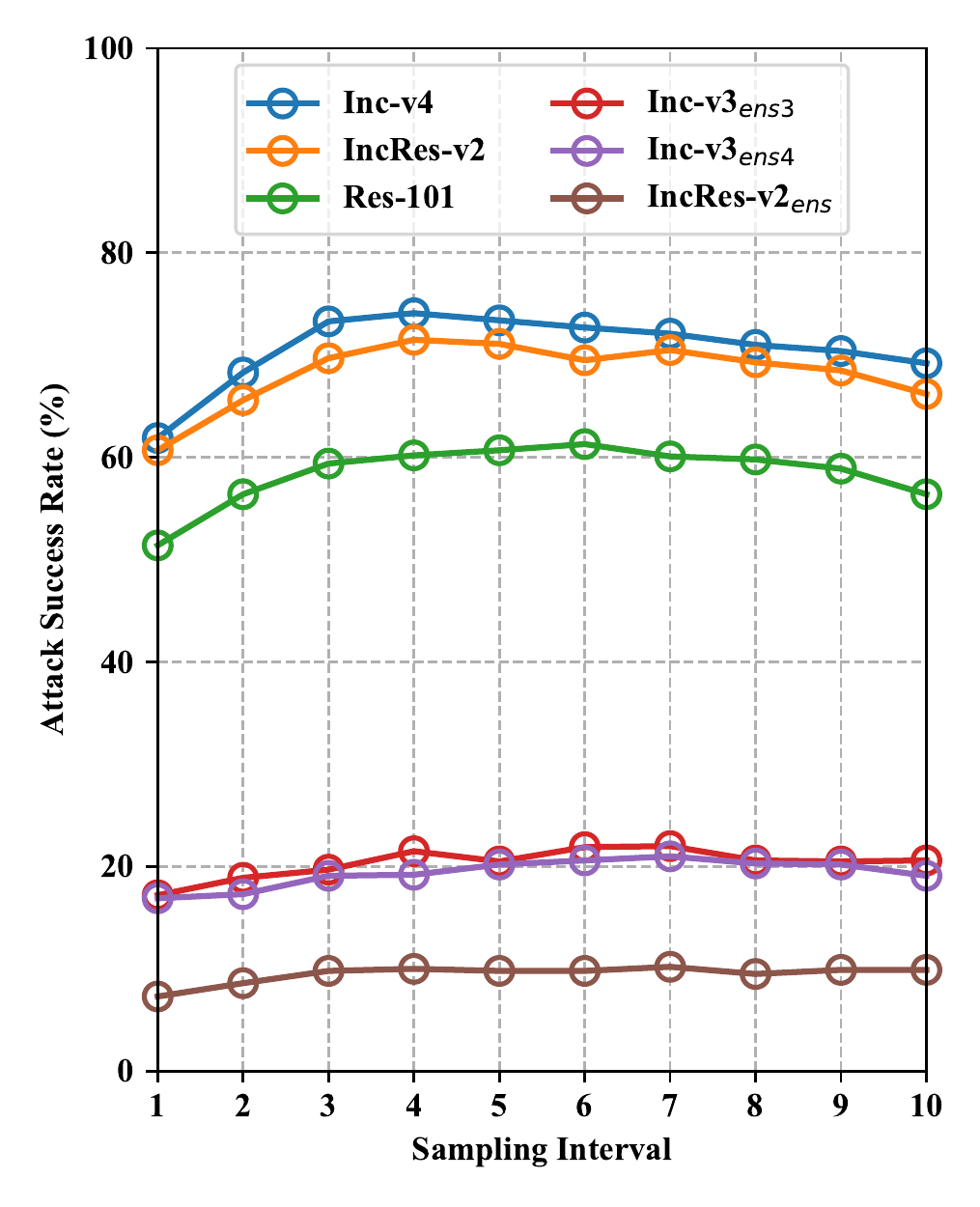}
          \caption{EMI-FGSM}
          \label{fig:sampleInterval:MLNI-FGSM}
        \end{subfigure}
        \hspace{.1em}
        \begin{subfigure}{.48\textwidth} 
          \centering 
          \includegraphics[width=\linewidth]{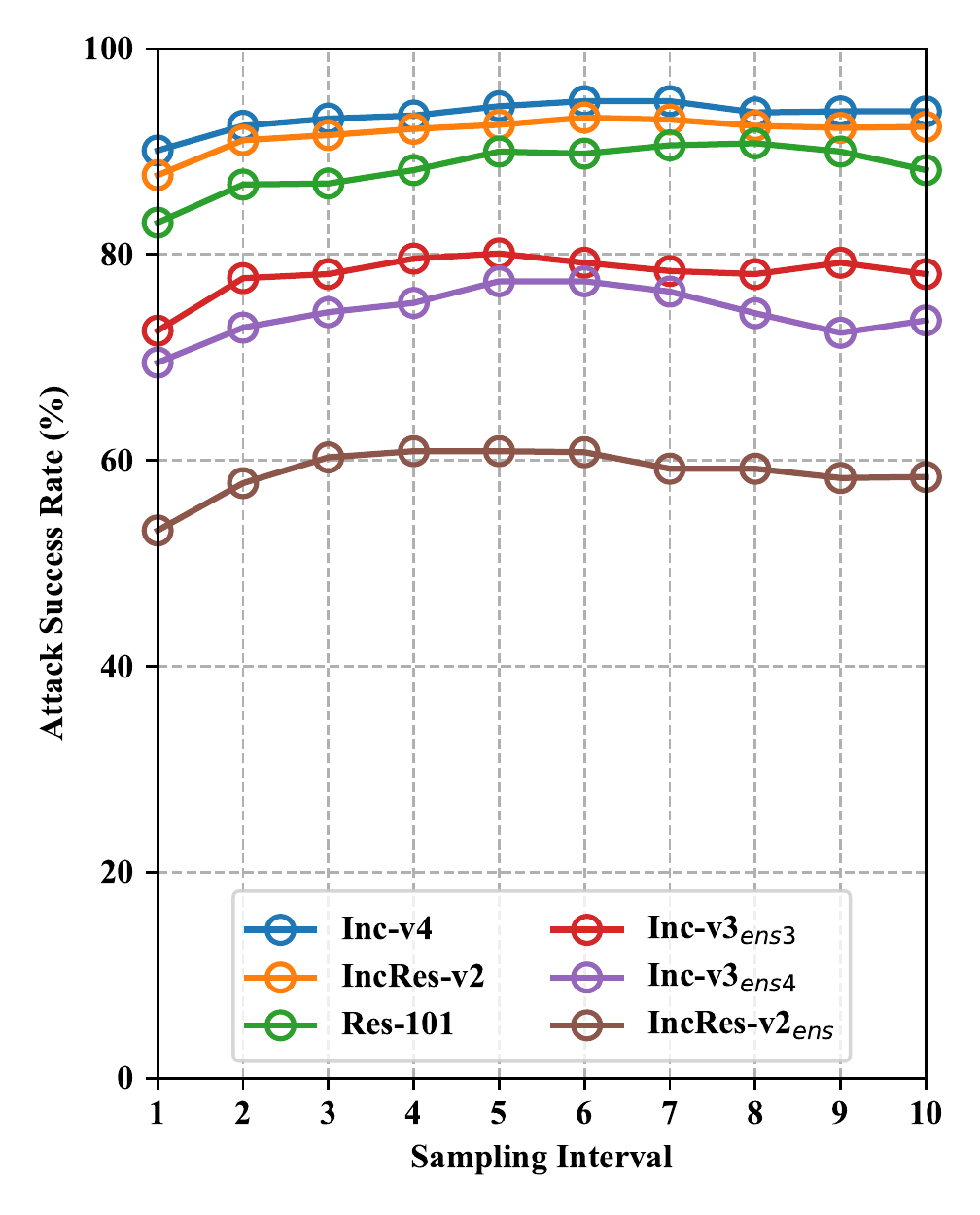}
          \caption{EMI-DTS}
          \label{fig:sampleInterval:MLNI-DTS}
        \end{subfigure}% 
    \vspace{-0.3em}
    \caption{Attack success rates (\%) on the other six models with adversarial examples generated by EMI-FGSM and EMI-DTS on Inc-v3 for various \textbf{sampling interval}.}
    \label{fig:sampleInterval}
    \end{minipage}
    \hspace{.3cm}
    \begin{minipage}[b]{0.48\textwidth} 
        \begin{subfigure}{.48\textwidth}
          \centering 
          \includegraphics[width=\linewidth]{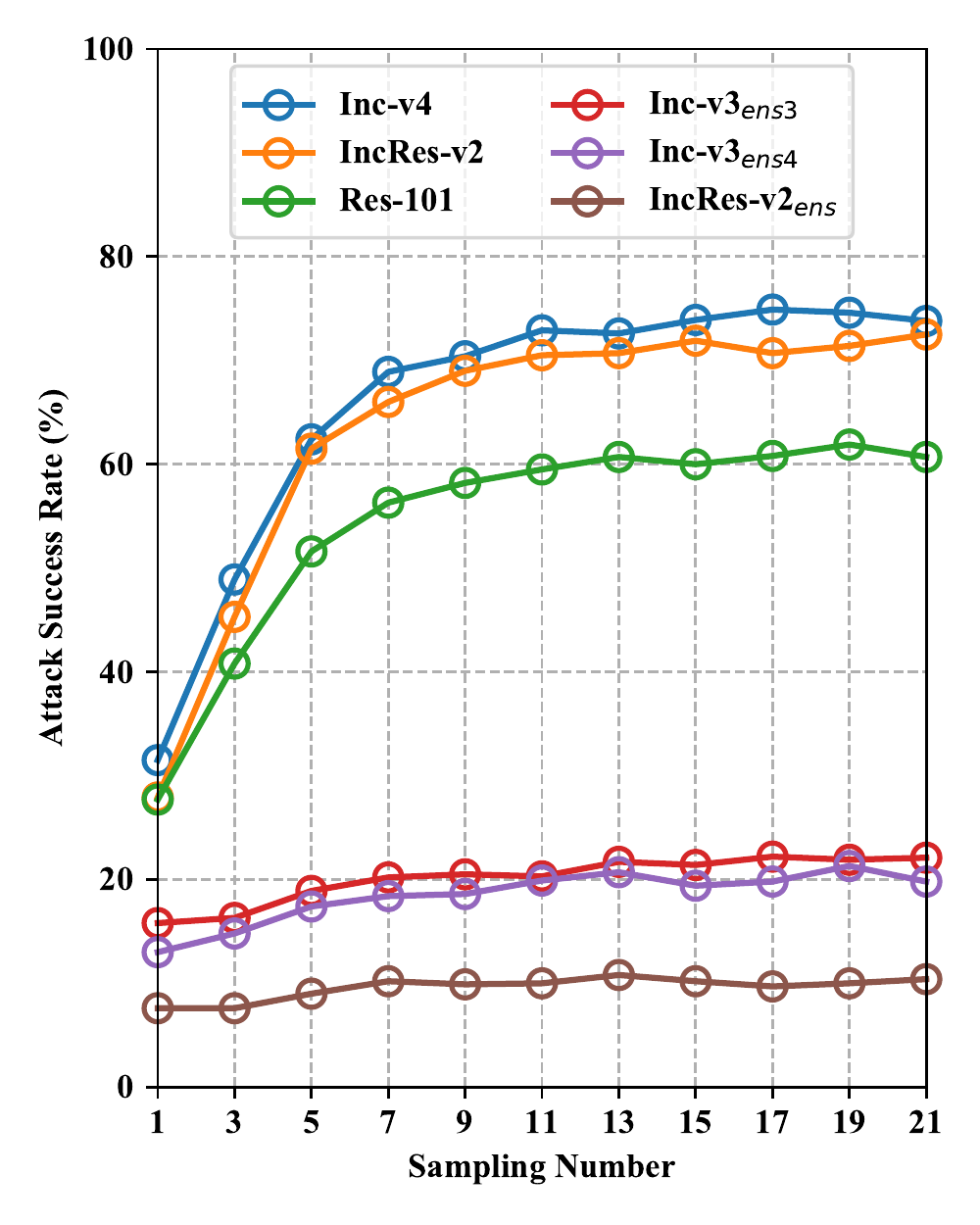}
          \caption{EMI-FGSM}
          \label{fig:sampleNumber:MLNI-FGSM}
        \end{subfigure}
        \hspace{.1em}
        \begin{subfigure}{.48\textwidth} 
          \centering 
          \includegraphics[width=\linewidth]{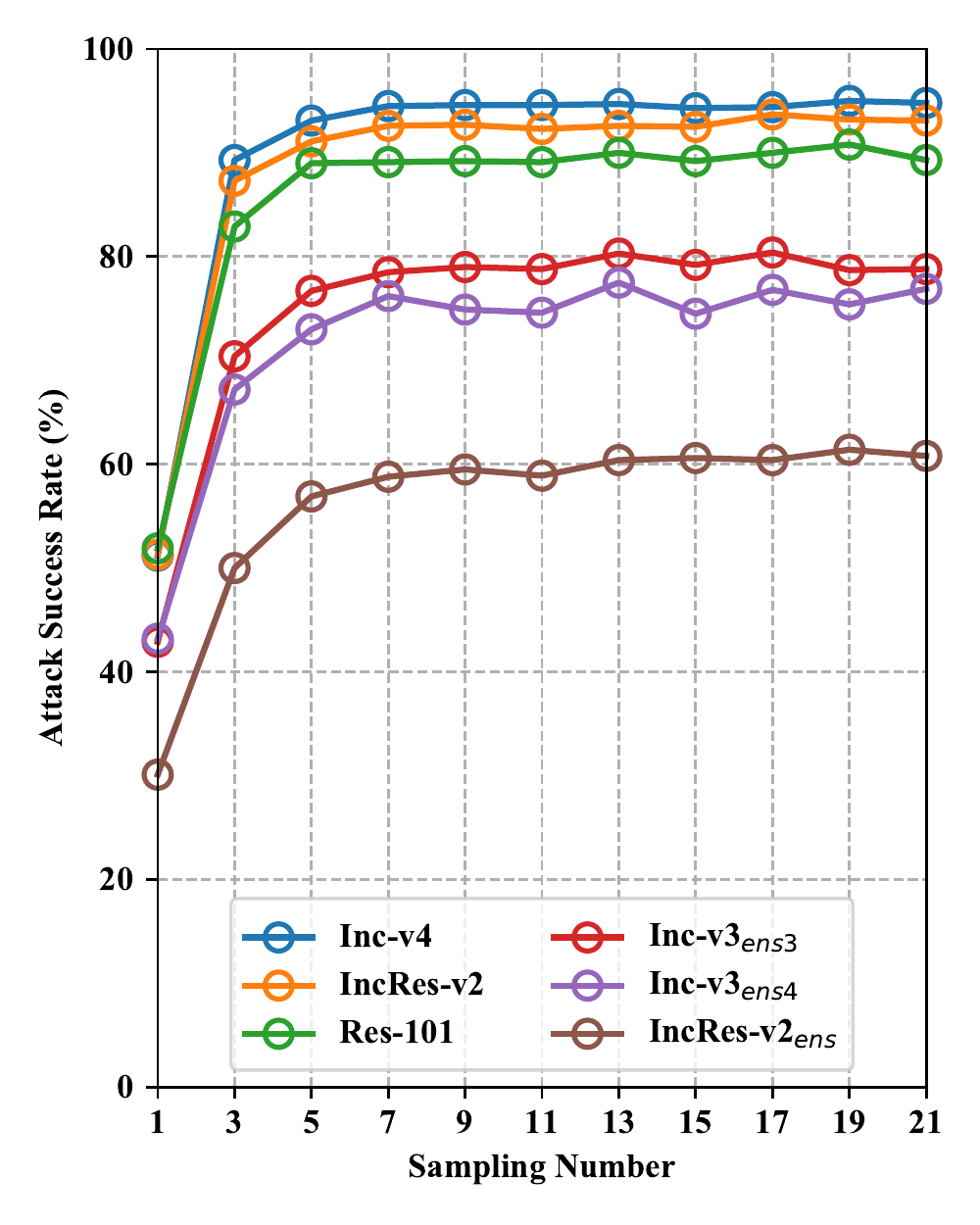}
          \caption{EMI-DTS}
          \label{fig:sampleNumber:MLNI-DTS}
        \end{subfigure}% 
        \vspace{-0.3em}
    \caption{Attack success rates (\%) on the other six models with adversarial examples generated by EMI-FGSM and EMI-DTS on Inc-v3 for various \textbf{sampling number}.}
    \label{fig:sampleNumber}
    \end{minipage}
    \vspace{-0.75em}
\end{figure*}

\textbf{On sampling interval.} The sampling interval $\eta$ also plays a key role in improving the attack performance. We try different values of $\eta$ from 1 to 10 and the results are summarized in Figure~\ref{fig:sampleInterval}. For all the values of $\eta$, the white-box attack success rate is 100\%. The transferability increases when $\eta \leq 3$ for both EMI-FGSM and EMI-DTS. For $4 \leq \eta \leq 7$, the attacks exhibit similar transferability and the performance decays slightly when $\eta > 7$. Thus we adopt $\eta=7$ in experiments.

% The transferability on normally trained models increases when $\eta$ increases and reaches the peak when $\eta=4$ for EMI-FGSM and $\eta=7$ for EMI-DTS. The transferability on adversarially trained models exhibits the same trend but reaches the peak when $\eta=7$ for EMI-FGSM and $\eta=5$ for EMI-DTS. In our experiments, we adopt $\eta=7$.
% \input{figs/ablation_sample_interval}

\textbf{On sampling number.} We continue to explore the impact of the sampling number $N$, as illustrated in Figure~\ref{fig:sampleNumber}. The white-box attack success rate for various values of $N$ is 100\%. When $N=1$, EMI-FGSM degrades to MI-FGSM and exhibits the lowest transferability. When we increase the value of $N$, the transferability increases rapidly before $N=11$ for EMI-FGSM and $N=7$ for EMI-DTS. When $N>11$, increasing $N$ can still bring small performance improvement for EMI-FGSM. However, the bigger the value of $N$, the higher the computational cost. To balance the performance gain and the cost, we set $N=11$ in experiments.

\begin{figure}
\centering 
        \begin{subfigure}{.23\textwidth}
          \centering 
          \includegraphics[width=\linewidth]{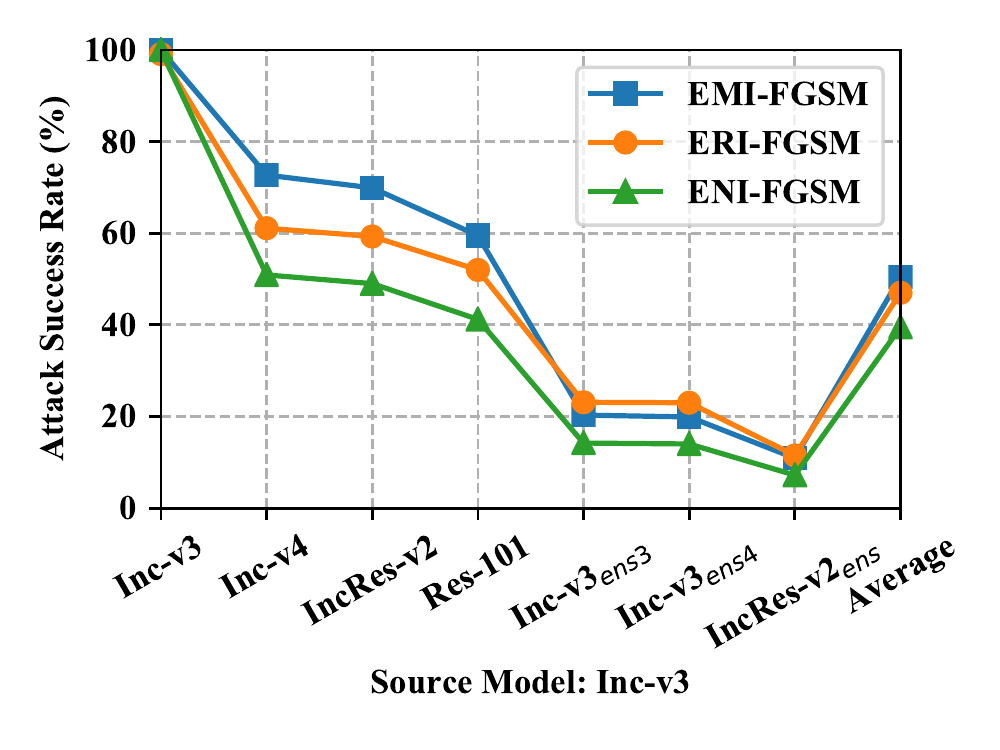}
        %   \vspace{-1.5em}
        \end{subfigure}
        % \hspace{.1em}
        \begin{subfigure}{.23\textwidth}
          \centering 
          \includegraphics[width=\linewidth]{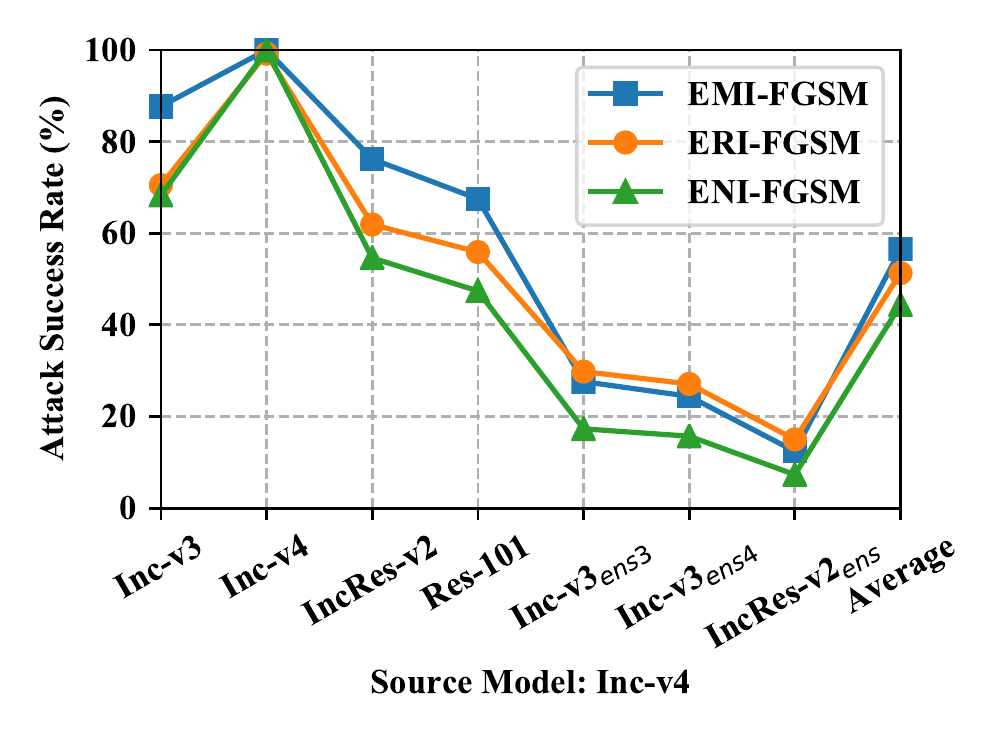}
        %   \vspace{-1.5em}
        \end{subfigure}\\
        
        \begin{subfigure}{.23\textwidth}
          \centering 
          \includegraphics[width=\linewidth]{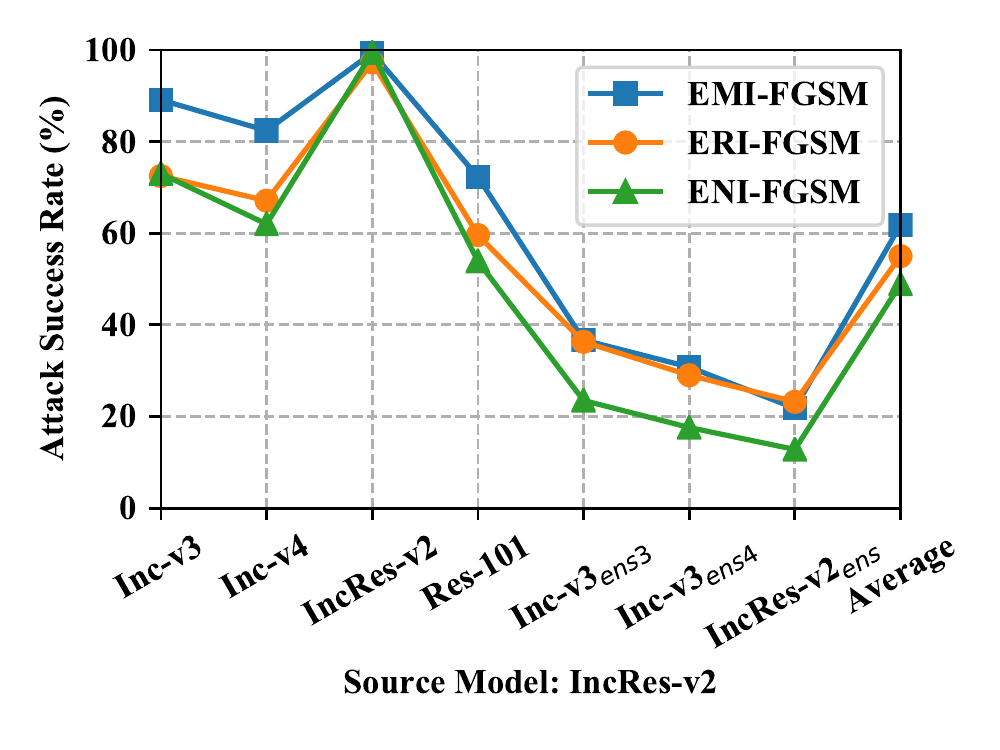}
        %   \vspace{-1.5em}
        \end{subfigure}
        % \hspace{.1em}
        \begin{subfigure}{.23\textwidth}
          \centering 
          \includegraphics[width=\linewidth]{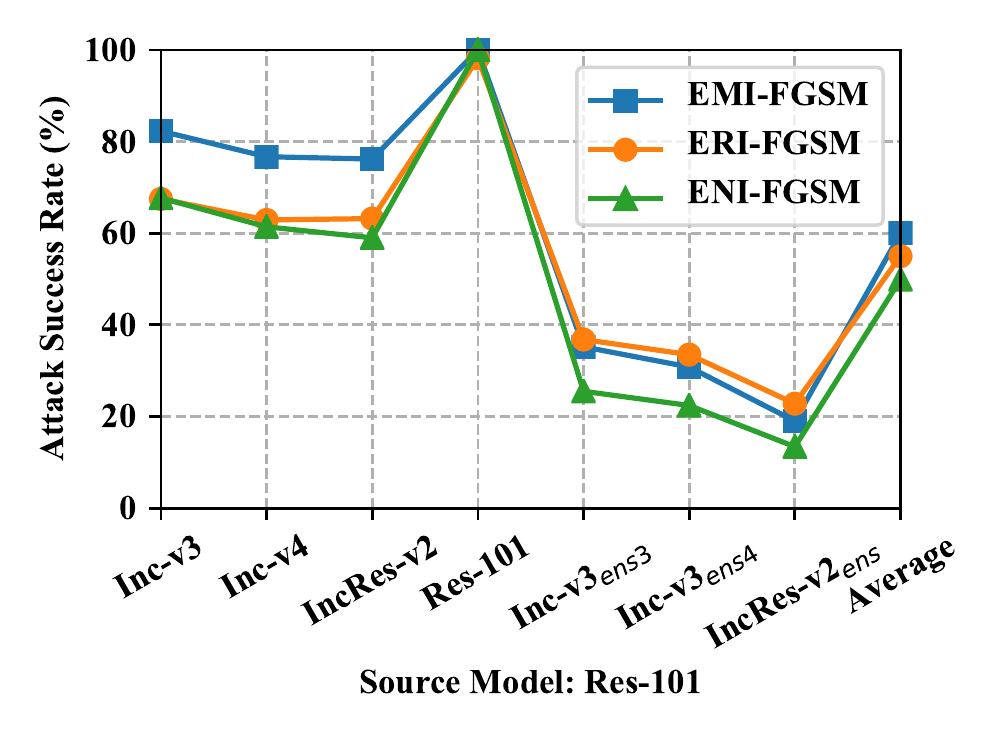}
        %   \vspace{-1.5em}
        \end{subfigure}
        % \hspace{.1em}
    \vspace{-0.5em}
    \caption{Attack success rates (\%) of EMI-FGSM, ERI-FGSM and ENI-FGSM against the seven models under single-model setting.}
    % \vspace{-0.5em}
    \label{fig:discussion_fig}
\end{figure}

% \begin{figure*}
% \centering 
%         \begin{subfigure}{.21\textwidth}
%           \centering 
%           \includegraphics[width=\linewidth]{figs/eps/discussion_inc_v3.eps}
%         %   \vspace{-1.5em}
%         \end{subfigure}
%         % \hspace{.1em}
%         \begin{subfigure}{.21\textwidth}
%           \centering 
%           \includegraphics[width=\linewidth]{figs/eps/discussion_inc_v4.eps}
%         %   \vspace{-1.5em}
%         \end{subfigure}
%         \begin{subfigure}{.21\textwidth}
%           \centering 
%           \includegraphics[width=\linewidth]{figs/eps/discussion_inc_res_v2.eps}
%         %   \vspace{-1.5em}
%         \end{subfigure}
%         % \hspace{.1em}
%         \begin{subfigure}{.21\textwidth}
%           \centering 
%           \includegraphics[width=\linewidth]{figs/eps/discussion_res_101.eps}
%         %   \vspace{-1.5em}
%         \end{subfigure}
%         % \hspace{.1em}
%     % \vspace{-0.5em}
%     \caption{Attack success rates (\%) of EMI-FGSM, ERI-FGSM and ENI-FGSM on seven models under single-model setting.}
%     % \vspace{-0.5em}
%     \label{fig:discussion_fig}
% \end{figure*}
\subsection{Discussion on Possible Variations}
%\subsection{Other Possible Momentum Enhancements}
\label{sec:discussion}
Except for EMI-FGSM, there are also other ways to enhance the momentum. Here we provide two possible implementations of the enhanced momentum, denoted as ENI-FGSM and ERI-FGSM. 
Specifically, ENI-FGSM samples the data points in the direction of momentum by substituting Eq. \ref{eq:sample_data} in EMI-FGSM with:
\begin{equation}
    \bar{x}_t^{adv}[i] = x_t^{adv} + c_i \cdot g_{t-1}
\end{equation}
where $g_{t-1}$ is the accumulated momentum of the previous iteration. ERI-FGSM adopts the accumulated gradient of randomly sampled data points by substituting Eq. \ref{eq:sample_data} in EMI-FGSM with:
\begin{equation}
    \bar{x}_t^{adv}[i] = x_t^{adv} + \alpha \cdot U(-1^d, 1^d)    
\end{equation}
where $U(a,b)$ denotes the uniform distribution in $[a,b]$.
% The details of ERI-FGSM and ENI-FGSM are provided in Appendix \ref{app:sec:variation}.
% To demonstrate the effectiveness of the proposed EMI-FGSM, we also test the performance of ENI-FGSM which adopts the accumulated gradient of the data points in the direction of momentum, and  ERI-FGSM which adopts the accumulated gradient of randomly sampled data points. The details of ERI-FGSM and ENI-FGSM are provided in Appendix \ref{app:sec:variation}.

\textbf{EMI-FGSM \vs ENI-FGSM.} The proposed EMI-FGSM accumulates the gradient of the data points in the direction of ($t$-1)-th gradient at the $t$-th iteration and exhibits remarkable performance improvement. However, from the perspective of NI-FGSM, \textit{can we accumulate the gradient of the data points in the direction of momentum at the $t$-th iteration?} To address this concern, we extend NI-FGSM to ENI-FGSM and test the attack performance. As shown in Figure~\ref{fig:discussion_fig}, we see that ENI-FGSM is considerably lower than EMI-FGSM. It further supports our hypothesis that the direction of the accumulated momentum cannot provide a precise description of the neighborhood and find proper point for the gradient calculation, as it contains too much accumulated information from the previous iterations. 

\textbf{EMI-FGSM \vs ERI-FGSM.} The comparison between ENI-FGSM and EMI-FGSM shows that the direction plays a big impact on the performance of the enhanced momentum. Both ENI-FGSM and EMI-FGSM sample the data points in a fixed direction. \textit{What if we accumulate the gradient of the data points in the neighborhood of $x_t^{adv}$ at the $t$-th iteration?} 
% To answer this question, we propose ERI-FGSM that adopts the accumulated gradient of the randomly sampled data points at each iteration and test the performance.
To address this concern, we test the performance of ERI-FGSM on various models.
As shown in Figure~\ref{fig:discussion_fig}, ERI-FGSM exhibits considerably lower transferabilty than EMI-FGSM on normally trained models but achieves slightly better performance on adversarially trained models. A possible reason might be that the data points with noise for the gradient calculation are more similar to the adversaries for the adversarial training.

\section{Conclusion}

Inspired by existing momentum based attacks, we propose an enhanced momentum method that not only accumulates the gradient of each iteration, but also accumulates the gradients of the sampled data points in the gradient direction of previous iteration. 
We then incorporate our enhanced momentum method into the iterative gradient-based methods to strengthen the adversarial attacks, which can significantly improve the attack success rates under white-box as well as black-box settings, 
as evaluated on the standard ImageNet dataset. 
Our strongest enhanced momentum  based attack, the EMI-DTS that is integrated with existing input transformations under the ensemble-model setting, could achieve an average black-box attack success rates of over 94\%, showing very high adversarial transferability.
Our work also indicates that existing defenses are far from being deployed in real-world applications and stronger robust deep learning models are needed.

%In this work, we propose to incorporate the enhanced momentum method into the iterative gradient-based methods to strengthen the adversarial attacks, which can effectively improve the attack success rates under white-box as well as black-box settings. Specifically, the enhanced momentum not only accumulates the gradient of each iteration, but also accumulates the gradients of the sampled data points in the gradient direction of previous iteration. Extensive experiments on the standard ImageNet dataset demonstrate that the enhanced momentum based attack can significantly boost the adversatial trasnferability. Moreover, integrating the proposed method with input transformations under ensemble-model setting achieves an average success rate of 86.6\% on various advanced defense models, outperforming the existing state-of-the-art attacks with a clear margin of 7.8\%. The results also demonstrate that existing defenses are far from being deployed in real-world applications and we need more robust deep learning models.

%From the optimization perspective, the enhanced momentum can be also regarded as a new optimization method for training to stabilize the training process and escape the local optima. We will adopt the enhanced momentum for the model training and provide theoretic analysis in our future work.

{\small
\bibliographystyle{ieee_fullname}
\bibliography{egbib}

\begin{thebibliography}{10}\itemsep=-1pt

\bibitem{athalye2018obfuscated}
Anish Athalye, Nicholas Carlini, and David Wagner.
\newblock Obfuscated gradients give a false sense of security: Circumventing
  defenses to adversarial examples.
\newblock {\em International Conference on Machine Learning (ICML)}, 2018.

\bibitem{carlini2017towards}
Nicholas Carlini and David Wagner.
\newblock Towards evaluating the robustness of neural networks.
\newblock In {\em 2017 IEEE Symposium on Security and Privacy (SP)}, pages
  39--57, 2017.

\bibitem{cohen2019certified}
Jeremy~M Cohen, Elan Rosenfeld, and J~Zico Kolter.
\newblock Certified adversarial robustness via randomized smoothing.
\newblock {\em International Conference on Machine Learning (ICML)}, 2019.

\bibitem{devlin2019bert}
Jacob Devlin, Ming-Wei Chang, Kenton Lee, and Kristina Toutanova.
\newblock Bert: Pre-training of deep bidirectional transformers for language
  understanding.
\newblock {\em Proceedings of the Conference of the North American Chapter of
  the Association for Computational Linguistics: Human Language Technologies
  (NAACL-HLT)}, 2019.

\bibitem{dong2018boosting}
Yinpeng Dong, Fangzhou Liao, Tianyu Pang, Hang Su, Jun Zhu, Xiaolin Hu, and
  Jianguo Li.
\newblock Boosting adversarial attacks with momentum.
\newblock In {\em Proceedings of the IEEE Conference on Computer Vision and
  Pattern Recognition (CVPR)}, pages 9185--9193, 2018.

\bibitem{dong2019evading}
Yinpeng Dong, Tianyu Pang, Hang Su, and Jun Zhu.
\newblock Evading defenses to transferable adversarial examples by
  translation-invariant attacks.
\newblock In {\em Proceedings of the IEEE Conference on Computer Vision and
  Pattern Recognition (CVPR)}, pages 4312--4321, 2019.

\bibitem{eykholt2018robust}
Kevin Eykholt, Ivan Evtimov, Earlence Fernandes, Bo Li, Amir Rahmati, Chaowei
  Xiao, Atul Prakash, Tadayoshi Kohno, and Dawn Song.
\newblock Robust physical-world attacks on deep learning visual classification.
\newblock In {\em Proceedings of the IEEE Conference on Computer Vision and
  Pattern Recognition (CVPR)}, pages 1625--1634, 2018.

\bibitem{girshick2015fast}
Ross Girshick.
\newblock Fast r-cnn.
\newblock In {\em Proceedings of the IEEE International Conference on Computer
  Vision (ICCV)}, pages 1440--1448, 2015.

\bibitem{goodfellow2015FGSM}
Ian~J Goodfellow, Jonathon Shlens, and Christian Szegedy.
\newblock Explaining and harnessing adversarial examples.
\newblock {\em International Conference on Learning Representations (ICLR)},
  2015.

\bibitem{he2016resnet}
Kaiming He, Xiangyu Zhang, Shaoqing Ren, and Jian Sun.
\newblock Deep residual learning for image recognition.
\newblock In {\em Proceedings of the IEEE conference on computer vision and
  pattern recognition (CVPR)}, pages 770--778, 2016.

\bibitem{krizhevsky2012imagenet}
Alex Krizhevsky, Ilya Sutskever, and Geoffrey~E Hinton.
\newblock Imagenet classification with deep convolutional neural networks.
\newblock In {\em Advances in Neural Information Processing Systems (NIPS)},
  pages 1097--1105, 2012.

\bibitem{kurakin2017IFGSM}
Alexey Kurakin, Ian Goodfellow, and Samy Bengio.
\newblock Adversarial examples in the physical world.
\newblock {\em International Conference on Learning Representations (ICLR),
  Workshop Track Proceedings}, 2017.

\bibitem{li2019nattack}
Yandong Li, Lijun Li, Liqiang Wang, Tong Zhang, and Boqing Gong.
\newblock Nattack: Learning the distributions of adversarial examples for an
  improved black-box attack on deep neural networks.
\newblock {\em International Conference on Machine Learning (ICML)}, 2019.

\bibitem{liao2018defense}
Fangzhou Liao, Ming Liang, Yinpeng Dong, Tianyu Pang, Xiaolin Hu, and Jun Zhu.
\newblock Defense against adversarial attacks using high-level representation
  guided denoiser.
\newblock In {\em Proceedings of the IEEE Conference on Computer Vision and
  Pattern Recognition (CVPR)}, pages 1778--1787, 2018.

\bibitem{lin2020nesterov}
Jiadong Lin, Chuanbiao Song, Kun He, Liwei Wang, and John~E Hopcroft.
\newblock Nesterov accelerated gradient and scale invariance for adversarial
  attacks.
\newblock In {\em International Conference on Learning Representations (ICLR)},
  2020.

\bibitem{liu2017delving}
Yanpei Liu, Xinyun Chen, Chang Liu, and Dawn Song.
\newblock Delving into transferable adversarial examples and black-box attacks.
\newblock {\em International Conference on Learning Representations (ICLR)},
  2017.

\bibitem{liu2019FD}
Zihao Liu, Qi Liu, Tao Liu, Nuo Xu, Xue Lin, Yanzhi Wang, and Wujie Wen.
\newblock Feature distillation: Dnn-oriented jpeg compression against
  adversarial examples.
\newblock In {\em Proceedings of the IEEE Conference on Computer Vision and
  Pattern Recognition (CVPR)}, pages 860--868, 2019.

\bibitem{long2015fully}
Jonathan Long, Evan Shelhamer, and Trevor Darrell.
\newblock Fully convolutional networks for semantic segmentation.
\newblock In {\em Proceedings of the IEEE conference on computer vision and
  pattern recognition (CVPR)}, pages 3431--3440, 2015.

\bibitem{madry2018pgd}
Aleksander Madry, Aleksandar Makelov, Ludwig Schmidt, Dimitris Tsipras, and
  Adrian Vladu.
\newblock Towards deep learning models resistant to adversarial attacks.
\newblock {\em International Conference on Learning Representations (ICLR)},
  2018.

\bibitem{Nesterov1983}
Yurii Nesterov.
\newblock A method for unconstrained convex minimization problem with the rate
  of convergence o(1/kˆ2).
\newblock {\em Doklady AN USSR}, 269:543--547, 1983.

\bibitem{papernot2017practical}
Nicolas Papernot, Patrick McDaniel, Ian Goodfellow, Somesh Jha, Z~Berkay Celik,
  and Ananthram Swami.
\newblock Practical black-box attacks against machine learning.
\newblock In {\em Proceedings of the 2017 ACM on Asia Conference on Computer
  and Communications Security}, pages 506--519, 2017.

\bibitem{polyak1964some}
Boris~T Polyak.
\newblock Some methods of speeding up the convergence of iteration methods.
\newblock {\em Ussr computational mathematics and mathematical physics},
  4(5):1--17, 1964.

\bibitem{russakovsky2015imagenet}
Olga Russakovsky, Jia Deng, Hao Su, Jonathan Krause, Sanjeev Satheesh, Sean Ma,
  Zhiheng Huang, Andrej Karpathy, Aditya Khosla, Michael Bernstein, et~al.
\newblock Imagenet large scale visual recognition challenge.
\newblock {\em International Journal of Computer Vision (IJCV)},
  115(3):211--252, 2015.

\bibitem{salman2019provably}
Hadi Salman, Jerry Li, Ilya Razenshteyn, Pengchuan Zhang, Huan Zhang, Sebastien
  Bubeck, and Greg Yang.
\newblock Provably robust deep learning via adversarially trained smoothed
  classifiers.
\newblock In {\em Advances in Neural Information Processing Systems (NeurIPS)},
  pages 11292--11303, 2019.

\bibitem{sharif2016accessorize}
Mahmood Sharif, Sruti Bhagavatula, Lujo Bauer, and Michael~K Reiter.
\newblock Accessorize to a crime: Real and stealthy attacks on state-of-the-art
  face recognition.
\newblock In {\em Proceedings of the 2016 ACM Sigsac Conference on Computer and
  Communications Security}, pages 1528--1540, 2016.

\bibitem{song2019improving}
Chuanbiao Song, Kun He, Liwei Wang, and John~E Hopcroft.
\newblock Improving the generalization of adversarial training with domain
  adaptation.
\newblock {\em International Conference on Learning Representations (ICLR)},
  2019.

\bibitem{szegedy2017inception}
Christian Szegedy, Sergey Ioffe, Vincent Vanhoucke, and Alex Alemi.
\newblock Inception-v4, inception-resnet and the impact of residual connections
  on learning.
\newblock {\em AAAI Conference on Artificial Intelligence (AAAI)}, 2017.

\bibitem{szegedy2016inceptionv3}
Christian Szegedy, Vincent Vanhoucke, Sergey Ioffe, Jon Shlens, and Zbigniew
  Wojna.
\newblock Rethinking the inception architecture for computer vision.
\newblock In {\em Proceedings of the IEEE conference on computer vision and
  pattern recognition (CVPR)}, pages 2818--2826, 2016.

\bibitem{szegedy2014intriguing}
Christian Szegedy, Wojciech Zaremba, Ilya Sutskever, Joan Bruna, Dumitru Erhan,
  Ian Goodfellow, and Rob Fergus.
\newblock Intriguing properties of neural networks.
\newblock {\em International Conference on Learning Representations (ICLR)},
  2014.

\bibitem{tramer2018ensemble}
Florian Tram{\`e}r, Alexey Kurakin, Nicolas Papernot, Ian Goodfellow, Dan
  Boneh, and Patrick McDaniel.
\newblock Ensemble adversarial training: Attacks and defenses.
\newblock {\em International Conference on Learning Representations (ICLR)},
  2018.

\bibitem{wang2021Admix}
Xiaosen Wang, Xuanran He, Jingdong Wang, and Kun He.
\newblock Admix: Enhancing the transferability of adversarial attacks.
\newblock {\em arXiv preprint arXiv:2102.00436}, 2021.

\bibitem{wang2020improving}
Yisen Wang, Difan Zou, Jinfeng Yi, James Bailey, Xingjun Ma, and Quanquan Gu.
\newblock Improving adversarial robustness requires revisiting misclassified
  examples.
\newblock In {\em International Conference on Learning Representations (ICLR)},
  2020.

\bibitem{xie2018mitigating}
Cihang Xie, Jianyu Wang, Zhishuai Zhang, Zhou Ren, and Alan Yuille.
\newblock Mitigating adversarial effects through randomization.
\newblock {\em International Conference on Learning Representations (ICLR)},
  2018.

\bibitem{xie2019improving}
Cihang Xie, Zhishuai Zhang, Yuyin Zhou, Song Bai, Jianyu Wang, Zhou Ren, and
  Alan~L Yuille.
\newblock Improving transferability of adversarial examples with input
  diversity.
\newblock In {\em Proceedings of the IEEE Conference on Computer Vision and
  Pattern Recognition (CVPR)}, pages 2730--2739, 2019.

\bibitem{xu2018BitReduction}
Weilin Xu, David Evans, and Yanjun Qi.
\newblock Feature squeezing: Detecting adversarial examples in deep neural
  networks.
\newblock {\em Network and Distributed System Security Symposium (NDSS)}, 2018.

\bibitem{zhang2019theoretically}
Hongyang Zhang, Yaodong Yu, Jiantao Jiao, Eric~P Xing, Laurent~El Ghaoui, and
  Michael~I Jordan.
\newblock Theoretically principled trade-off between robustness and accuracy.
\newblock {\em International Conference on Machine Learning (ICML)}, 2019.

\end{thebibliography}
}

\clearpage
\appendix
{\centering
\section*{Appendix}}

In the supplementary material, we report the comparison results on various gradient-based attack methods and  our EMI-FGSM method integrated with various transformation-based methods when attacking the other three normally trained models, \ie Inc-v4, IncRes-v2, and Res-101 respectively. 

We first report the attack success rates of various gradient-based attack methods on the other three normally-trained models. The results are summarized in Table \ref{tab:comparison_three_models}. Compared with other advanced attacks, EMI-FGSM also exhibits better white-box attack success rates and higher transferability, which are consistent to the results on the Inc-v3 model in the main text.

% The results for various attacks under ensemble-model setting, where the adversarial examples are crafted on the ensemble of Inc-v3, Inc-v4, IncRes-v2, and Res-101, are depicted in Table \ref{tab:comparison_ens}. It can be observed that our proposed PI-FGSM and EMI-FGSM exhibit higher transferbility than the baseline attacks. Meanwhile, EMI-FGSM achieves the highest attack success rates among all the attack methods, which further validates the effectiveness of the proposed method.

The results for the EMI-FGSM integrated with various transformation-based methods under single-model setting, where the adversarial examples are crafted on the other three normally-trained models, are depicted in Table \ref{tab:transformation_three_models}.  It can be observed that EMI significantly promotes the attack success rates of the baseline attacks with a clear margin, which are consistent to the results on the Inc-v3 model in the main text and further verifies the high effectiveness of the proposed enhanced momentum.

\newpage

\begin{table*}[b]
% \vspace{-9.5em}
\begin{center}
\begin{subtable}{1.\textwidth}
\begin{center}
%\resizebox{\textwidth}{!}{
 \scalebox{0.9}{
\begin{tabular}{l|cccccccc} 
\hline
Attack & Inc-v3 & Inc-v4* & IncRes-v2 & Res-101 & Inc-v3${\rm _{ens3}}$ & Inc-v3${\rm _{ens4}}$ & IncRes-v2${\rm _{ens}}$ \\ 
\hline \hline
FGSM & 27.4 & ~~52.0 & 22.5 & 22.9 & 15.7 & ~~9.4 & ~~5.4 \\
I-FGSM & 32.8 & \textbf{100.0} & 20.0 & 19.9 & ~~5.3 & ~~6.8 & ~~3.1 \\
MI-FGSM & 56.2 & ~~99.9 & 46.0 & 40.7 & 15.7 & 15.1 & ~~8.3  \\
NI-FGSM & 63.0 & ~~99.9 & 52.4 & 45.6 & 16.5 & 14.3 & ~~7.5  \\
% RI-FGSM & 55.8 & 99.9 & 45.5 & 41.2 & 17.2 & 15.3 & 7.6 & 40.4 \\
\hline
PI-FGSM (\textbf{Ours}) & 72.4 & ~~99.9 & 59.7 & 52.5 & 18.0 & 15.7 & ~~7.3  \\
% ENI-FGSM & 68.3 & 99.9 & 54.6 & 47.4 & 17.3 & 15.7 & 7.3 & 44.4 \\
% ERI-FGSM & 70.5 & 99.2 & 61.9 & 55.9 & \textbf{29.8} & \textbf{27.1} & \textbf{15.0} & 51.3 \\
EMI-FGSM (\textbf{Ours}) & \textbf{87.7} & \textbf{100.0} & \textbf{76.2} & \textbf{67.5} & \textbf{27.6} & \textbf{24.4} & \textbf{12.4} \\
\hline
\end{tabular}
 }
\caption{Attack success rates (\%) for the adversarial examples crafted on \textbf{Inc-v4}.}
\end{center}
\end{subtable}
\end{center}
\vspace{-1em}
\begin{center}
\begin{subtable}{1.\textwidth}
\begin{center}
%\resizebox{\textwidth}{!}{
 \scalebox{0.9}{
\begin{tabular}{l|cccccccc} 
\hline
Attack & Inc-v3 & Inc-v4 & IncRes-v2* & Res-101 & Inc-v3${\rm _{ens3}}$ & Inc-v3${\rm _{ens4}}$ & IncRes-v2${\rm _{ens}}$ \\ 
\hline \hline
FGSM & 27.2 & 20.2 & 41.9 & 23.6 & ~~9.5 & ~~9.1 & ~~5.7 \\
I-FGSM & 33.4 & 25.2 & 98.2 & 20.2 & ~~6.8 & ~~6.4 & ~~4.3 \\  
MI-FGSM & 57.3 & 50.4 & 98.2 & 44.7 & 21.2 & 16.0 & 11.5 \\
NI-FGSM & 63.4 & 55.9 & 99.0 & 45.3 & 20.2 & 15.8 & 10.0 \\
% RI-FGSM & 59.8 & 51.5 & 97.4 & 44.4 & 21.1 & 16.5 & 11.2 & 43.1 \\
\hline
PI-FGSM (\textbf{Ours}) & 71.6 & 63.4 & 98.3 & 53.4 & 24.3 & 18.7 & 12.5 \\
% ENI-FGSM & 73.0 & 62.0 & \textbf{99.4} & 53.9 & 23.5 & 17.6 & 12.8 & 48.9 \\
% ERI-FGSM & 72.5 & 67.1 & 97.3 & 59.6 & 36.3 & 29.0 & 23.2 & 55.0 \\
EMI-FGSM (\textbf{Ours}) & \textbf{89.1} & \textbf{82.4} & \textbf{99.4} & \textbf{72.3} & \textbf{36.7} & \textbf{30.8} & \textbf{21.8} \\
\hline
\end{tabular}
 }
\caption{Attack success rates (\%) for the adversarial examples crafted on \textbf{IncRes-v2}.}
\end{center}
\end{subtable}
\end{center}
\vspace{-1em}
\begin{center}
\begin{subtable}{1.\textwidth}
\begin{center}
%\resizebox{\textwidth}{!}{
 \scalebox{0.9}{
\begin{tabular}{l|cccccccc} 
\hline
Attack & Inc-v3 & Inc-v4 & IncRes-v2 & Res-101* & Inc-v3${\rm _{ens3}}$ & Inc-v3${\rm _{ens4}}$ & IncRes-v2${\rm _{ens}}$ \\ 
\hline \hline
FGSM & 36.4 & 31.2 & 30.0 & ~~78.1 & 14.9 & 13.3 & ~~6.5 \\
I-FGSM & 31.4 & 25.3 & 23.1 & ~~99.3 & ~~8.7 & ~~8.5 & ~~5.4 \\  
MI-FGSM & 57.6 & 51.9 & 49.8 & ~~99.3 & 23.9 & 22.1 & 12.6 \\
NI-FGSM & 65.5 & 58.0 & 57.5 & ~~99.4 & 24.3 & 21.5 & 11.3 \\
% RI-FGSM & 57.4 & 51.6 & 49.6 & 99.3 & 24.3 & 21.8 & 12.1 & 45.2 \\
\hline
PI-FGSM (\textbf{Ours}) & 72.8 & 66.8 & 63.7 & ~~99.3 & 28.3 & 25.3 & 14.0 \\
% ENI-FGSM & 67.7 & 61.4 & 59.0 & \textbf{100.0} & 25.5 & 22.4 & 13.4 & 49.9 \\
% ERI-FGSM & 67.5 & 62.9 & 63.2 & 98.2 & \textbf{36.8} & \textbf{33.5} & \textbf{22.8} & 55.0 \\
EMI-FGSM (\textbf{Ours}) & \textbf{82.3} & \textbf{76.7} & \textbf{76.2} & \textbf{100.0} & \textbf{35.2} & \textbf{30.8} & \textbf{19.0} \\
\hline
\end{tabular}
 }
\caption{Attack success rates (\%) for the adversarial examples crafted on \textbf{Res-101}.}
\end{center}
\end{subtable}
\caption{\textbf{Attack success rates (\%) of various adversarial attacks against the seven baseline models under single-model setting.} The adversarial examples are crafted on \textbf{Inc-v4, IncRes-v2 or Res-101} using various adversarial attack methods. * indicates the white-box model being attacked.}
\label{tab:comparison_three_models}
\end{center}
\end{table*}

\begin{table*}[b]
% \vspace{-9.5em}
\begin{center}
\begin{subtable}{1.\textwidth}
\begin{center}
% \resizebox{\textwidth}{!}{
 \scalebox{0.9}{
\begin{tabular}{l|cccccccc} 
\hline
Attack & Inc-v3 & Inc-v4* & IncRes-v2 & Res-101 & Inc-v3${\rm _{ens3}}$ & Inc-v3${\rm _{ens4}}$ & IncRes-v2${\rm _{ens}}$\\ 
\hline \hline
DIM & 74.1 & ~~98.5 & 66.3 & 58.0 & 22.3 & 21.0 & 11.6 \\
EMI-DIM (\textbf{Ours}) & \textbf{89.4} & \textbf{~~99.1} & \textbf{83.6} & \textbf{75.2} & \textbf{33.5} & \textbf{30.9} & \textbf{16.7} \\
\hline
TIM & 58.0 & ~~99.5 & 47.2 & 42.8 & 25.9 & 24.0 & 16.9 \\
EMI-TIM (\textbf{Ours}) & \textbf{89.0} & \textbf{~~99.8} & \textbf{81.2} & \textbf{72.3} & \textbf{52.1} & \textbf{48.3} & \textbf{35.2} \\
\hline
SIM & 80.6 & ~~99.5 & 73.6 & 68.8 & 47.9 & 44.9 & 29.2 \\
EMI-SIM (\textbf{Ours}) & \textbf{96.4} & \textbf{~~99.9} & \textbf{93.7} & \textbf{89.0} & \textbf{59.7} & \textbf{56.1} & \textbf{36.9} \\
\hline
DTS & 84.7 & ~~98.0 & 80.5 & 76.3 & 67.9 & 66.9 & 54.3 \\
EMI-DTS (\textbf{Ours}) & \textbf{95.7} & \textbf{~~99.4} & \textbf{94.5} & \textbf{90.7} & \textbf{81.4} & \textbf{77.5} & \textbf{68.8} \\
\hline
\end{tabular}
 }
\caption{Attack success rates (\%) for the adversarial examples crafted on \textbf{Inc-v4}.}
\end{center}
\end{subtable}
\end{center}
\vspace{-1em}
\begin{center}
\begin{subtable}{1.\textwidth}
\begin{center}
% \resizebox{\textwidth}{!}{
 \scalebox{0.9}{
\begin{tabular}{l|cccccccc} 
\hline
Attack & Inc-v3 & Inc-v4 & IncRes-v2* & Res-101 & Inc-v3${\rm _{ens3}}$ & Inc-v3${\rm _{ens4}}$ & IncRes-v2${\rm _{ens}}$ \\ 
\hline \hline
DIM & 68.1 & 65.1 & ~~93.7 & 58.3 & 30.2 & 23.4 & 17.3 \\
EMI-DIM (\textbf{Ours}) & \textbf{88.8} & \textbf{85.1} & \textbf{~~98.5} & \textbf{78.3} & \textbf{42.4} & \textbf{35.5} & \textbf{26.4} \\
\hline
TIM & 62.1 & 55.8 & ~~97.2 & 49.9 & 31.0 & 28.3 & 21.5 \\
EMI-TIM (\textbf{Ours}) & \textbf{90.6} & \textbf{85.0} & \textbf{~~99.4} & \textbf{80.1} & \textbf{61.5} & \textbf{52.2} & \textbf{48.2} \\
\hline
SIM & 84.6 & 79.5 & ~~98.9 & 76.1 & 55.9 & 49.0 & 41.7 \\
EMI-SIM (\textbf{Ours}) & \textbf{97.5} & \textbf{95.1} & \textbf{~~99.9} & \textbf{90.9} & \textbf{69.0} & \textbf{60.1} & \textbf{51.6} \\
\hline
DTS & 87.1 & 84.3 & ~~96.6 & 81.4 & 76.4 & 73.3 & 69.4 \\
EMI-DTS (\textbf{Ours}) & \textbf{97.8} & \textbf{95.4} & \textbf{~~99.9} & \textbf{93.6} & \textbf{88.2} & \textbf{83.5} & \textbf{81.9} \\
\hline
\end{tabular}
 }
\caption{Attack success rates (\%) for the adversarial examples crafted on \textbf{IncRes-v2}.}
\end{center}
\end{subtable}
\end{center}
\vspace{-1em}
\begin{center}
\begin{subtable}{1.\textwidth}
\begin{center}
% \resizebox{\textwidth}{!}{
 \scalebox{0.9}{
\begin{tabular}{l|cccccccc} 
\hline
Attack & Inc-v3 & Inc-v4 & IncRes-v2 & Res-101* & Inc-v3${\rm _{ens3}}$ & Inc-v3${\rm _{ens4}}$ & IncRes-v2${\rm _{ens}}$ \\ 
\hline \hline
DIM & 73.6 & 68.5 & 69.5 & ~~97.6 & 36.2 & 31.9 & 20.6 \\
EMI-DIM (\textbf{Ours}) & \textbf{88.7} & \textbf{84.3} & \textbf{84.1} & \textbf{~~99.7} & \textbf{46.4} & \textbf{40.7} & \textbf{26.3} \\
\hline
TIM & 59.4 & 54.0 & 52.3 & ~~99.2 & 35.6 & 31.8 & 22.8 \\
EMI-TIM (\textbf{Ours}) & \textbf{86.0} & \textbf{79.1} & \textbf{79.8} & \textbf{100.0} & \textbf{56.4} & \textbf{50.3} & \textbf{41.7} \\
\hline
SIM & 74.4 & 69.8 & 68.3 & ~~99.7 & 43.1 & 39.4 & 26.0 \\
EMI-SIM (\textbf{Ours}) & \textbf{92.0} & \textbf{88.7} & \textbf{88.4} & \textbf{100.0} & \textbf{57.6} & \textbf{50.4} & \textbf{35.7} \\
\hline
DTS & 84.0 & 80.0 & 81.9 & ~~98.9 & 73.3 & 70.9 & 59.3 \\
EMI-DTS (\textbf{Ours}) & \textbf{93.7} & \textbf{90.9} & \textbf{92.3} & \textbf{~~99.6} & \textbf{83.9} & \textbf{80.9} & \textbf{71.7} \\
\hline
\end{tabular}
 }
\caption{Attack success rates (\%) for the adversarial examples crafted on \textbf{Res-101}.}
\end{center}
\end{subtable}
\caption{\textbf{Attack success rates (\%) of various adversarial attacks against the seven baseline models under single-model setting.} The adversarial examples are crafted on \textbf{Inc-v4, IncRes-v2 or Res-101} using various adversarial attack methods. * indicates the white-box model being attacked.}
\label{tab:transformation_three_models}
\end{center}
% \vspace{6em}
\end{table*}

\end{document}